\documentclass[a4paper,conference]{IEEEtran}

\ifCLASSINFOpdf
\else
\fi

\usepackage{times}
\usepackage{epsfig}
\usepackage{graphicx}
\usepackage{amsmath}
\usepackage{amssymb}
\usepackage{enumitem}
\usepackage{subcaption}
\usepackage{multirow}
\usepackage{booktabs}
\usepackage{xcolor}
\usepackage{amssymb}
\usepackage{blindtext}
\usepackage{pifont}
\newcommand{\xmark}{\ding{55}}

\usepackage[pagebackref=true,breaklinks=true,letterpaper=true,colorlinks,urlcolor=blue,bookmarks=false]{hyperref}

\title{Tilting at windmills: Data augmentation for deep pose estimation does not help with occlusions}
\author{\IEEEauthorblockN{Rafal Pytel}
\IEEEauthorblockA{Computer Vision Lab\\
Delft University of Technology\\
}
\and
\IEEEauthorblockN{Osman Semih Kayhan}
\IEEEauthorblockA{Computer Vision Lab\\
Delft University of Technology
}
\and
\IEEEauthorblockN{Jan C. van Gemert}
\IEEEauthorblockA{Computer Vision Lab\\
Delft University of Technology 
}}
\begin{document}

\maketitle

\begin{abstract}

Occlusion degrades the performance of human pose estimation. In this paper, we introduce targeted keypoint and body part occlusion attacks. The effects of the attacks are systematically analyzed on the best performing methods. In addition, we propose occlusion specific data augmentation techniques against keypoint and part attacks. 
Our extensive experiments show that human pose estimation methods are not robust to occlusion and data augmentation does not solve the occlusion problems.
\footnote{For the code and the extended version: 

\url{https://github.com/rpytel1/occlusion-vs-data-augmentations}}
\end{abstract}

\IEEEpeerreviewmaketitle

\section{Introduction}


Human Pose Estimation is the task of localizing anatomical keypoints such as eyes, hips, knees and localizing  body-parts like  head, limbs, corpus, etc., with many applications in  segmentation~\cite{ li2019pose2body, lin2020cross,Xia_2017_CVPR}, action recognition~\cite{Luvizon_2020,Nie2015JointAR, Soomro2019OnlineLA}, pose tracking~\cite{girdhar2018detecttrack, xiu2018poseflow}, gait recognition~\cite{pose_gait, soft_biometrics}, autonomous driving~{\cite{crossing, cluenet, limb_detection}}, elderly monitoring~\cite{dias2020gaze, obdrvzalek2012accuracy} and social behaviour analysis~\cite{joint_picto,varadarajan2018joint}. All these applications rely on correct and robust pose estimation. In this paper we investigate the robustness of human pose estimation methods  to a natural and common effect: Occlusions.



Occlusions are common and occur frequently in the wild as for example  by a random object, another person~\cite{real_world_crowd}, and self-occlusion \cite{followmeup}. Prior works address occlusion in a general way and exploits segmentation~\cite{cluenet} or depth information~\cite{Semantic_occl}. Where~\cite{hpe_robustness} evaluates robustness with image and domain-agnostic universal perturbations.
In contrast, we systematically analyze  targeted occlusion attacks not only for keypoints, but also for and body parts and investigate the sensitivity of pose estimation to occlusion attacks.


A promising solution to occlusions is data augmentation, which is practically a default setting for deep learning applications~\cite{shorten2019survey} where image flipping, rotation, and scaling offer endless data variations~\cite{higher_hrnet, shorten2019survey, taylor2017improving}. As such, regional dropout and mixup methods improve the generalization performance of image classification~\cite{cutout, mixup_manifold,  mixed_mixup, mixup_between, manifold, holistic_cutout, mixup, random_erase}, object localization and detection~\cite{attention_dropout, dvornik2018modeling, hide_seek} and segmentation~\cite{cutmix_semantic}. In pose estimation, \cite{keypoint_cutmix} applies region based augmentation by exchanging a single keypoint patch with a random background patch. More recent approaches \cite{ HRNet, simple-baseline} use half-body augmentation wherewith the presence of more than 8 keypoints, by choosing upper or lower body keypoints. 
We implement  systematic data augmentation methods for occlusion for keypoint and body parts to investigate how  data augmentation can remedy occlusion attacks.


We have the following contributions: First, we conduct a structured investigation on the occlusion problem of pose estimation and introduce occlusion attacks. Second, we investigate occlusion-based data augmentation methods. Third, we show that data augmentation does not provide robustness to occlusion attacks.

\begin{figure}
\centering
  \begin{subfigure}{\columnwidth}
  \includegraphics[width=0.95\textwidth]{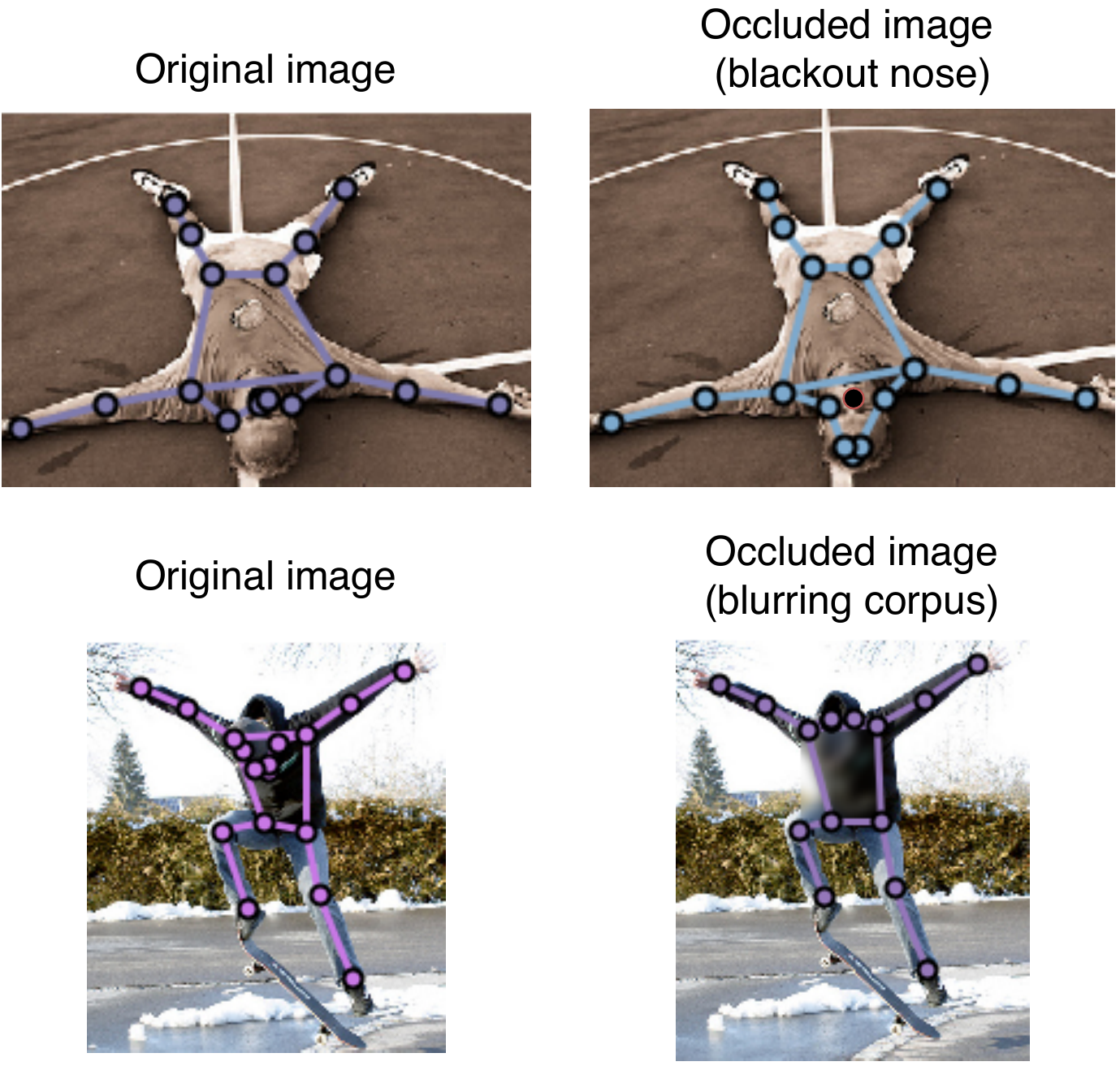} 
  \end{subfigure}
  \caption{Qualitative example how HRNet-32~\cite{HRNet} predictions change after keypoint blackout on the nose (first row) and part blurring on the corpus (second row). For both examples keypoints change for head, nose, eyes and ears. }
  \label{fig:examples}
  \end{figure}

\section{Related work}

 \begin{figure}
  \begin{subfigure}{0.32\columnwidth}
  \includegraphics[width=\textwidth]{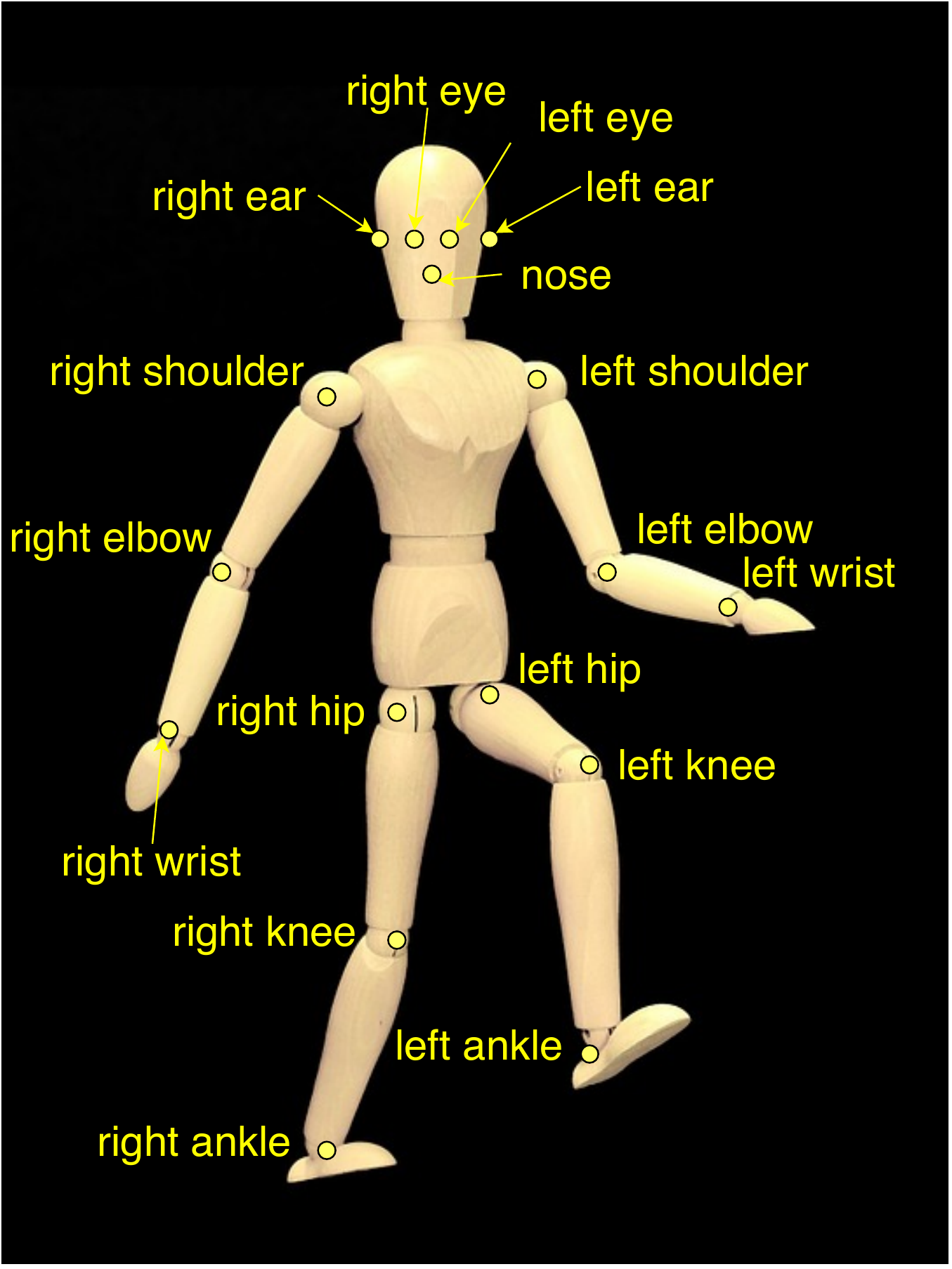}
  \caption{COCO keypoints annotations.}
  \end{subfigure}
  \begin{subfigure}{0.32\columnwidth}
  \includegraphics[width=\textwidth]{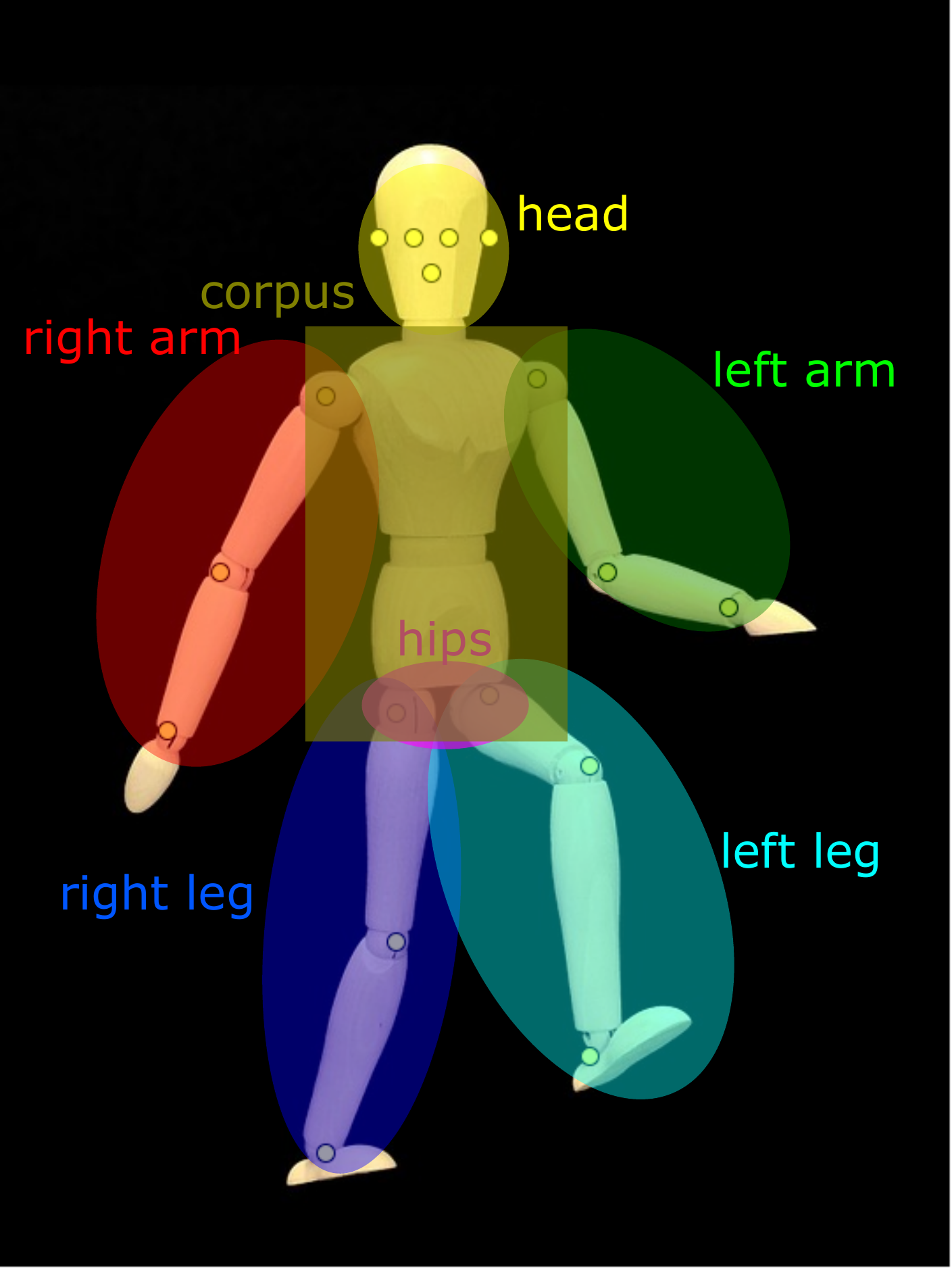}
  \caption{Part mapping for smaller parts.}
  \end{subfigure} 
  \begin{subfigure}{0.32\columnwidth} 
  \includegraphics[width=\textwidth]{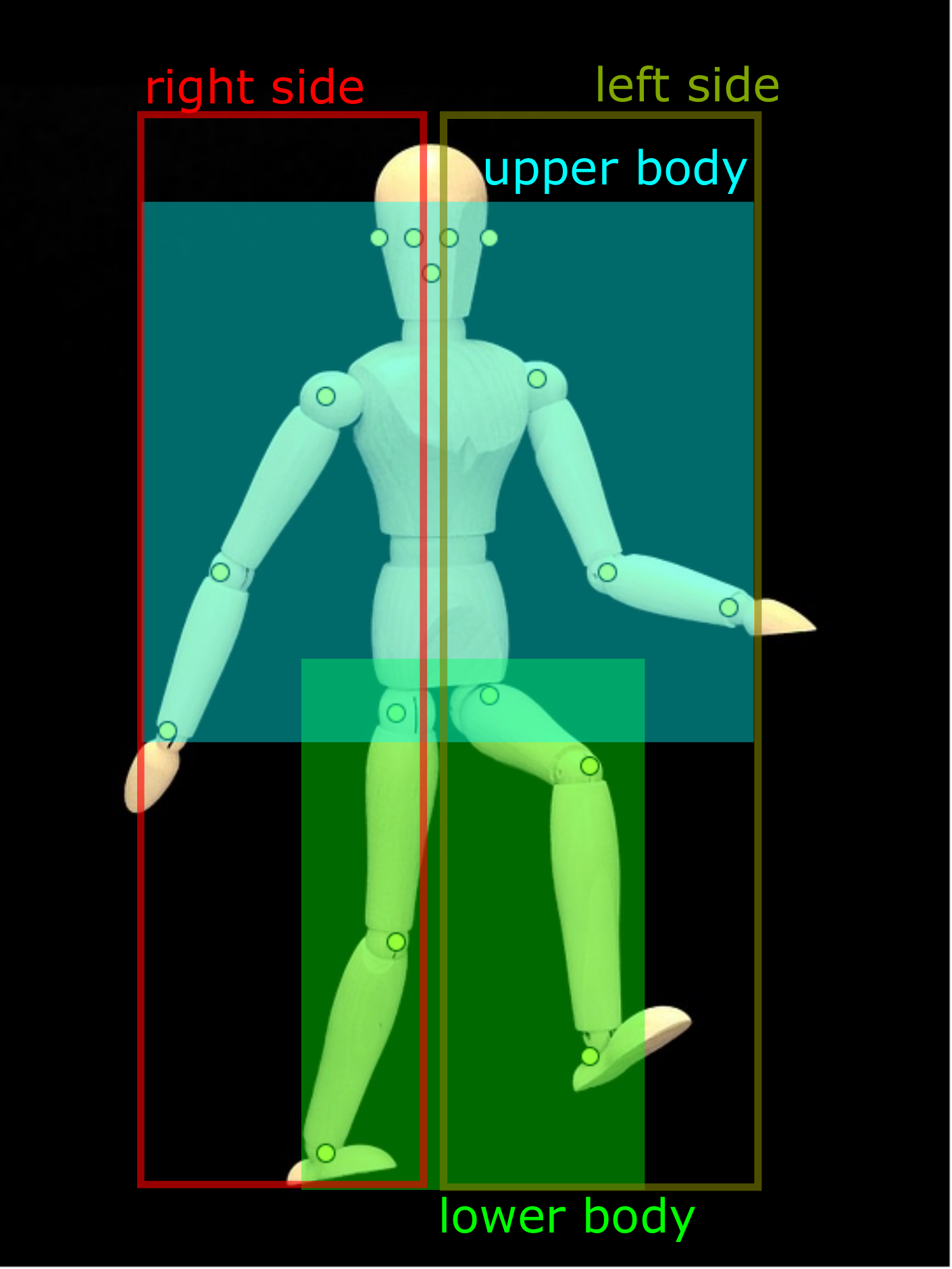} 
  \caption{Part mapping for larger parts.}
  \end{subfigure}  
\caption{Visualization of keypoint annotations  in COCO dataset and proposed part mapping. }
\label{fig:woodman}
\end{figure}

\textbf{Human Pose Estimation.} Deep learning methods in human pose estimation can be divided into 2 categories: bottom-up and top-down. Bottom-up approaches \cite{openpose, higher_hrnet, pifpaf}, firstly localize identity-free keypoints and then group them into person instances. Top-down approaches \cite{cpn, hourglass,HRNet, simple-baseline} firstly detect a person in the image and then perform a single person estimation within the bounding box. 
The top-down approaches achieve the state of the art results on various multi-person benchmarks such as COCO \cite{COCO_dataset}, MPII \cite{mpii}. Within top-down approaches 2 categories can be distinguished: regressing direct location of each keypoint \cite{iterative,deep_pose} and keypoint heatmaps estimation \cite{hpe_attention, hourglass, HRNet, cp_machines, simple-baseline} followed by choosing the locations with the highest heat values as the keypoints. The best performing methods on COCO keypoint challenge use a cascade network \cite{cpn,Li2019RethinkingOM} to improve keypoint prediction. The 'SimpleBaseline'~\cite{simple-baseline} proposes simple but effective improvement by adding few deconvolutional layers to enlarge the resolution of output features. 
HRNet~\cite{HRNet} which is built from multiple branches can produce high-resolution feature maps with rich semantics and performs well on COCO. Some works advance performance of HRNet via improvement over standard encoding and decoding of heatmaps \cite{dark} and basing data processing on the unit length instead of pixels \cite{udp_hrnet} with an additional off-set strategy for encoding and decoding. 
Because of their good accuracy and wide adaptation, we focus on top-down methods, HRNet and SimpleBaseline and bottom-up approach Higher HRNet.

\textbf{Occlusion in pose estimation.} 
Occlusion in pose estimation is an under-studied problem. In~\cite{hpe_robustness} analyses of occlusions are done for deep pose estimators by domain-agnostic universal perturbations.
More recently, attempts to solve the occlusion problem in pose estimation are suggested via the usage of segmentation of occluded parts \cite{cluenet} and depth of in an image \cite{Semantic_occl}.
OcclusionNet~\cite{occlusionet} predicts occluded keypoints via graph-neural networks yet it is applied only on vehicles.
Different from these methods, in our paper we introduce  keypoint occlusion attacks and body part occlusion attacks and give a structured analysis of occlusion on human pose estimation.

\textbf{ Data augmentation.} Data augmentation is a strong, simple and popular approach to increase model robustness. Removing part of the image improves generalization of image classification \cite{cutout, holistic_cutout, random_erase} and object localization-detection \cite{attention_dropout, dvornik2018modeling, hide_seek}. 
Mixup \cite{ mixup_manifold,mixup_between, mixup} approaches which create a combination of two images are often used in image classification.
\cite{cutmix_semantic}\cite{cutmix} combine regional dropout and MixUp methods for image segmentation \cite{cutmix_semantic} and image classification \cite{cutmix} task.
\cite{keypoint_cutmix} proposes a cutmix-like approach where a small patch from the background is pasted on the single keypoint or vice versa.
For the human pose estimation methods \cite{iterative,cp_machines, learning_pyramid}, scaling, rotation and flipping is commonly used as data augmentation. Random cropping is also used in bottom-up approaches \cite{openpose,higher_hrnet, pifpaf}. More recent top-down approaches \cite{cpn,HRNet, simple-baseline} employ the usage of half body transform
by a probability of 0.3 choosing either upper or lower body keypoints. 
We introduce and evaluate new data augmentation methods for keypoint and for body parts specifically designed against occlusion attacks for human pose estimation.

\begin{figure}
\centering
\subcaptionbox{}{\includegraphics[width=.49\columnwidth]{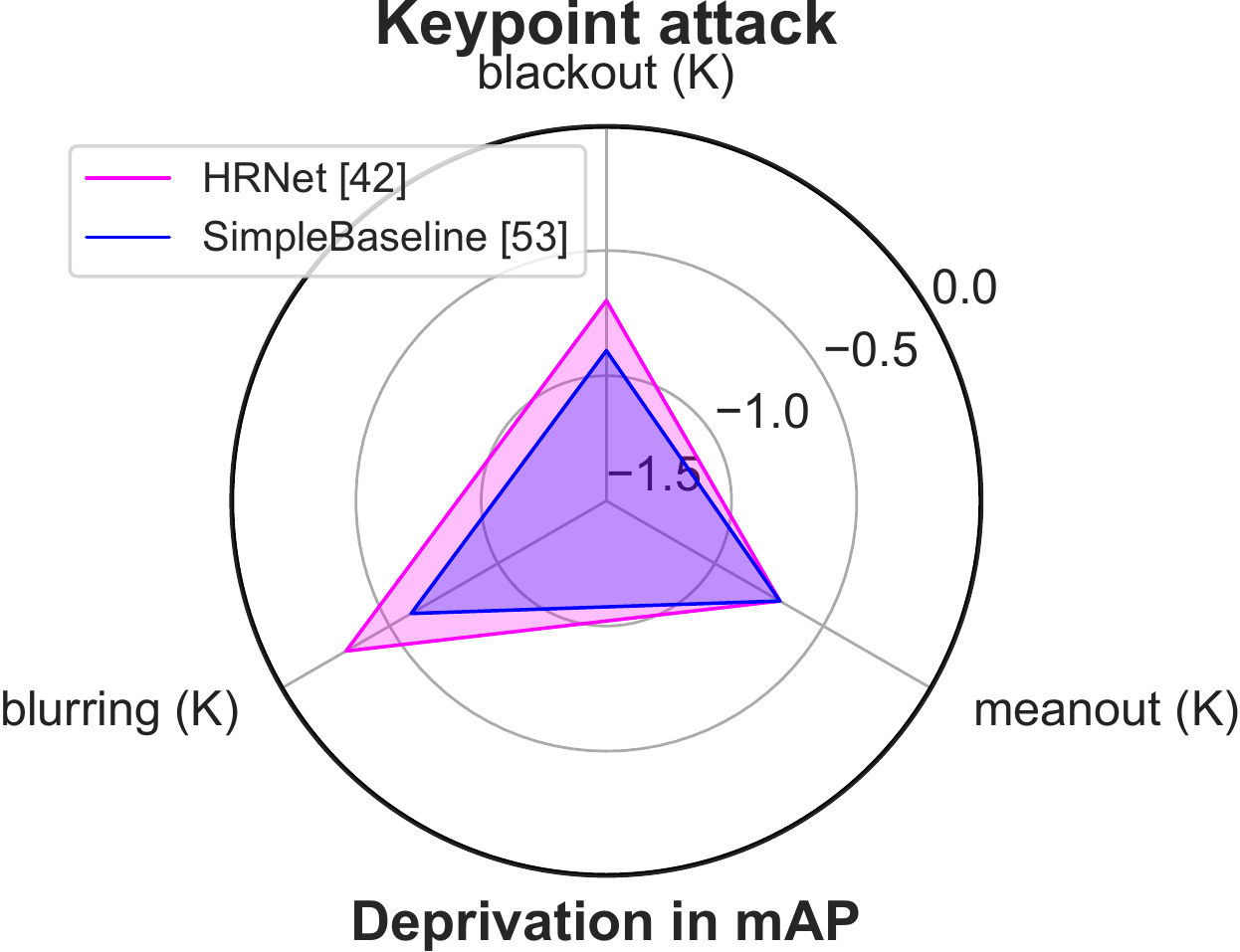}}
\hfill
\subcaptionbox{}{\includegraphics[width=.49\columnwidth]{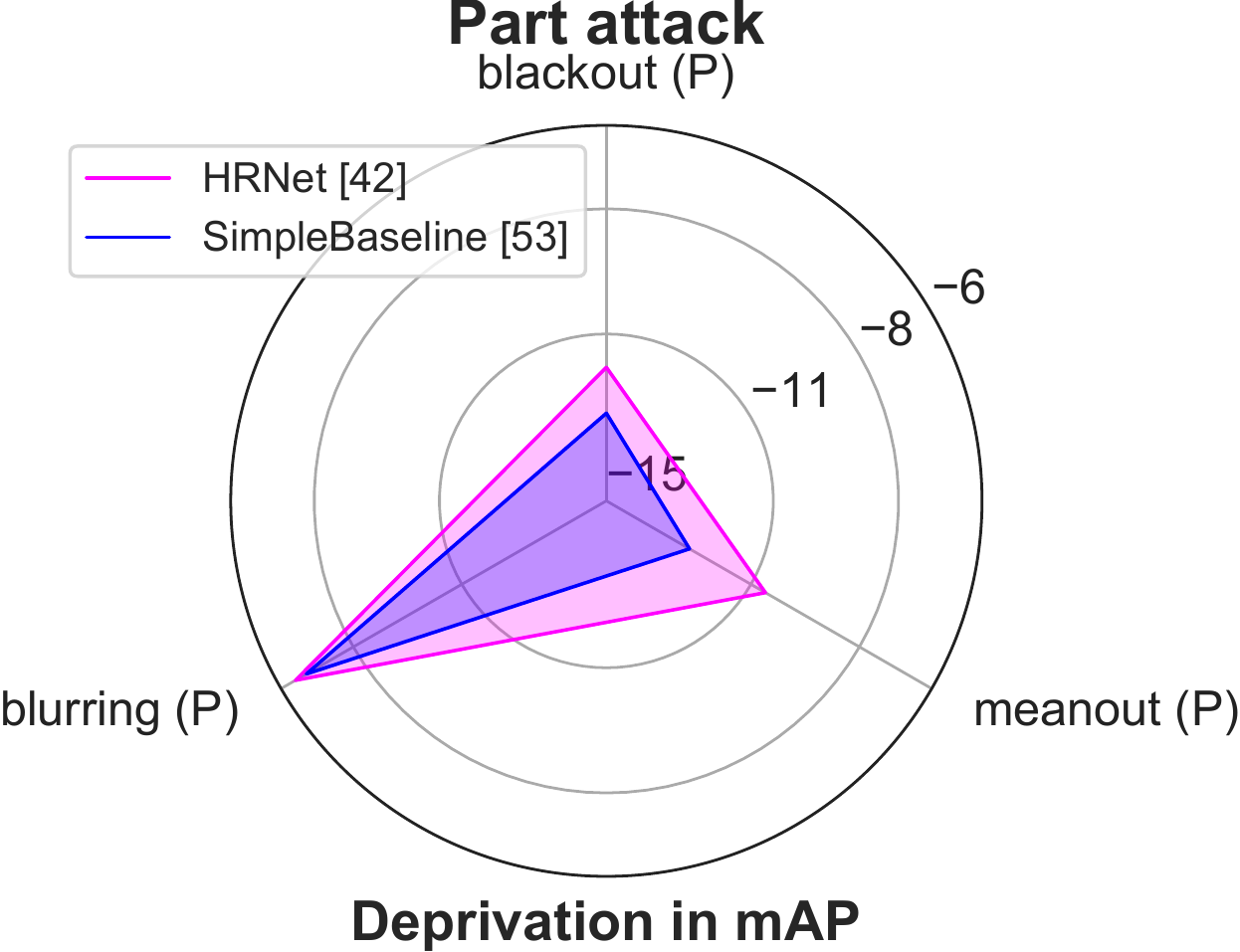}}
\caption{
Robustness comparison of HRNet \cite{HRNet} and SimpleBaseline \cite{simple-baseline} against (a) keypoint and (b) part occlusion attacks. HRNet is more robust against both attacks, yet both attacks drop performance, where part attacks deteriorate more.}
\label{hr_vs_simple}
\end{figure}

\section{Sensitivity to occlusion attacks}


We investigate the effect of occlusion attacks on MS COCO dataset~\cite{COCO_dataset}. COCO contains challenging images with the unconstrained environment, different body scales, variety of human poses and occlusion patterns.
The dataset contains over 200k images with 250k person instances labelled with 17 keypoints. Models are trained on COCO train2017 datasets which includes 57k images and 150k person instances. 
The evaluation is done on val2017 set which contains 5k images.

The occlusion attack experiments are conducted with HRNet \cite{HRNet} and Simple Baseline \cite{simple-baseline} for two  aspects: (i) keypoint attacks, where the occlusion area is a centred circle  on the chosen keypoint, (ii) body part attacks, where the occlusion area is the minimum rectangle covering all keypoints of a chosen part. 
The COCO keypoints and the proposed groups of body parts can be seen in Figure~\ref{fig:woodman}. For the analyses, COCO pretrained HRNet and Simple Baseline are evaluated by the performance of the network against keypoint and part occlusion attacks on COCO validation set. 

HRNet and SimpleBaseline produce heatmap instead of predicting direct single location for each keypoint.
The ground truth heatmaps are generated by using 2D Gaussian of size 13x13. Thus, as a default, we choose the size of the occlusion circle with a radius of 6 pixels for keypoint attacks to cover the keypoint heatmap. We have 3 different keypoint attacks: (i) Gaussian blur (blurring) attack, (ii) attack by filling with black pixels (blackout), (iii) attack by filling with a mean intensity value of a given image (meanout).

Body parts occlusion attacks are designed to draw a minimum rectangle which covers all the keypoints of a chosen part. Similar to the keypoint attacks, we have 3 different part attacks which are applied to the occlusion area: blurring with the kernel size 31 and sigma 5, blackout and meanout. 
These attacks can be applied on both small parts such as head, arms, hips and larger parts like upper body, lower body, left and right side (Figure~\ref{fig:woodman} b and c).

We compare HRNet and Simple Baseline according to their robustness to keypoint and part occlusion attacks. Figure~\ref{hr_vs_simple} shows that both attacks are quite successful as occlusion causes the performance to drop. HRNet is more robust against keypoint and part occlusion attacks.  For further analyses, we only use HRNet as a baseline for our investigations. 

\subsection{How sensitive to key point occlusion attacks?}



\begin{figure}
    \centering
    \includegraphics[width = 0.85\columnwidth]{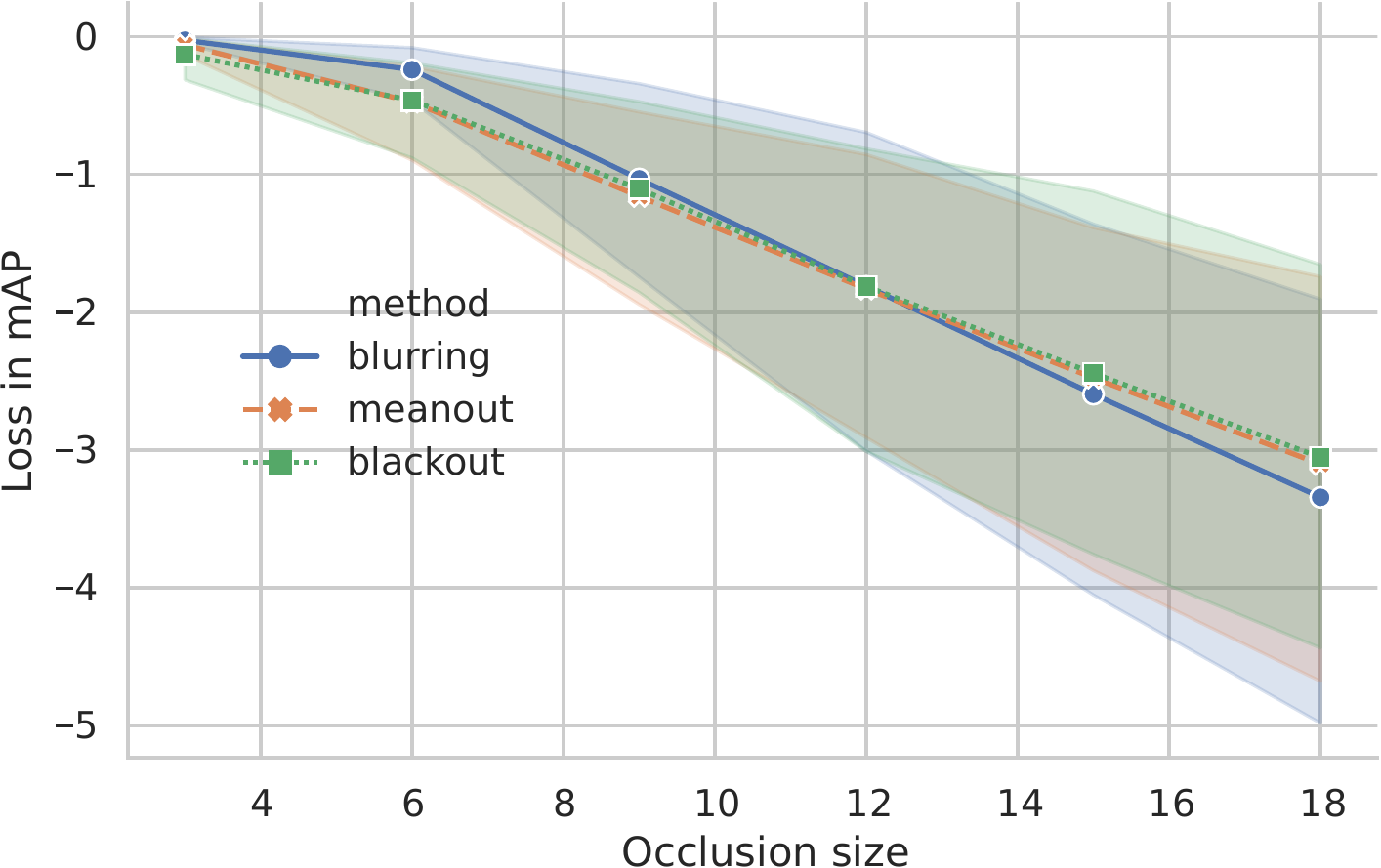}
    \caption{The relation between occlusion size and average loss in performance for keypoint level methods. Occlusion size greatly affects the performance.}
    \label{fig:patch}
\end{figure}

First, we analyze the effect of the occlusion size on the average performance of the pose estimator on all keypoints. Figure~\ref{fig:patch} indicates that pose estimator performance is inversely proportional to the occlusion size and blurring, blackout, and meanout attacks on average perform similarly. The size of the occlusion decreases the average performance of the estimator by approximately $3\%$ when the radius of the occlusion circle is chosen as 18 pixels. 

\begin{figure*}[!ht]
    \centering
    \includegraphics[height = 0.125 \textheight]{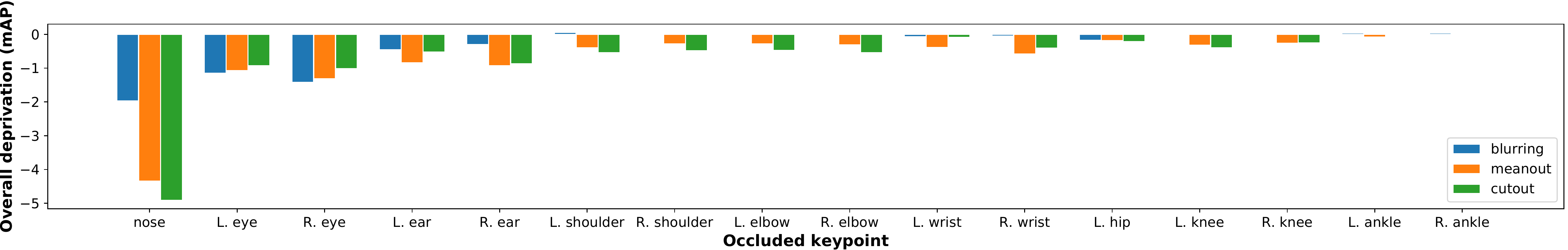}
    \caption{Overall loss in mAP after performing keypoint level occlusion. \textit{L.} and \textit{R.} correspond to the left and right side respectively. To note that, the occluded keypoint is included in the evaluation. Occluding nose causes the highest loss in performance.}
    \label{fig:keypoint-occlusions}
\end{figure*}

Second, we show the class-specific performance drops for each individual keypoints for each attack. In Figure \ref{fig:keypoint-occlusions},
attacking nose causes serious loss in mAP, almost $5\%$ for blackout, $4.4\%$ for meanout and $1.2\%$ for blurring.
The empirical results indicate that \textbf{the nose} is the most important keypoint since the occlusion of the nose causes notable performance drop. 
After the nose, each eye influences the performance of other keypoints mostly by approximately $1\%$ with each occlusion attack.
Keypoints from less densely annotated places like ankles or wrists are the least influential.

\begin{figure*}[ht!]
\centering
\subcaptionbox{The nose is the most influential keypoint causes a significant drop in the performance for the closest keypoints - left eye and right eye by around $10\%$.\label{fig:keypoint-occlusions-nose}
}{\includegraphics[height=.125\textheight]{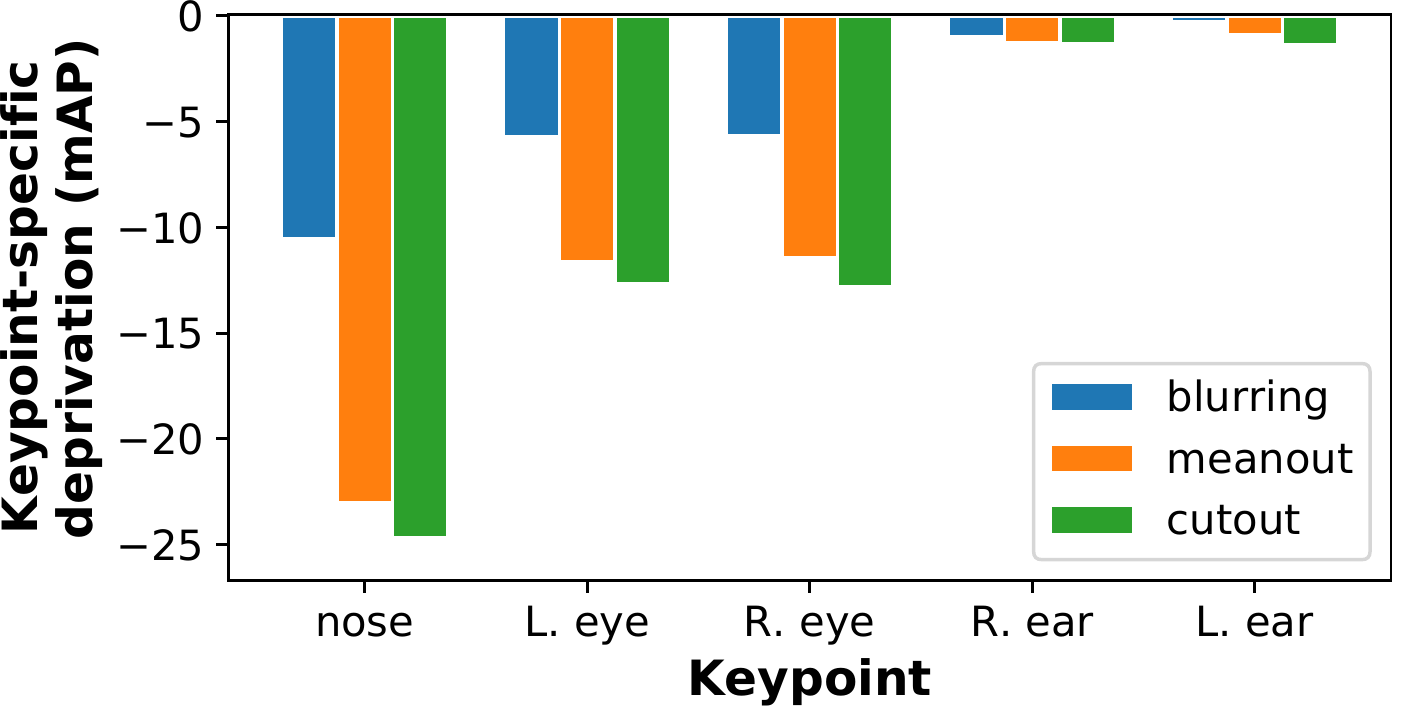}}
\hfill
\subcaptionbox{ When we occlude the left eye, there is a smaller loss in keypoint-specific performance for the left eye than while occluding nose.    \label{fig:keypoint-occlusions-left-eye}
}{\includegraphics[height=.125\textheight]{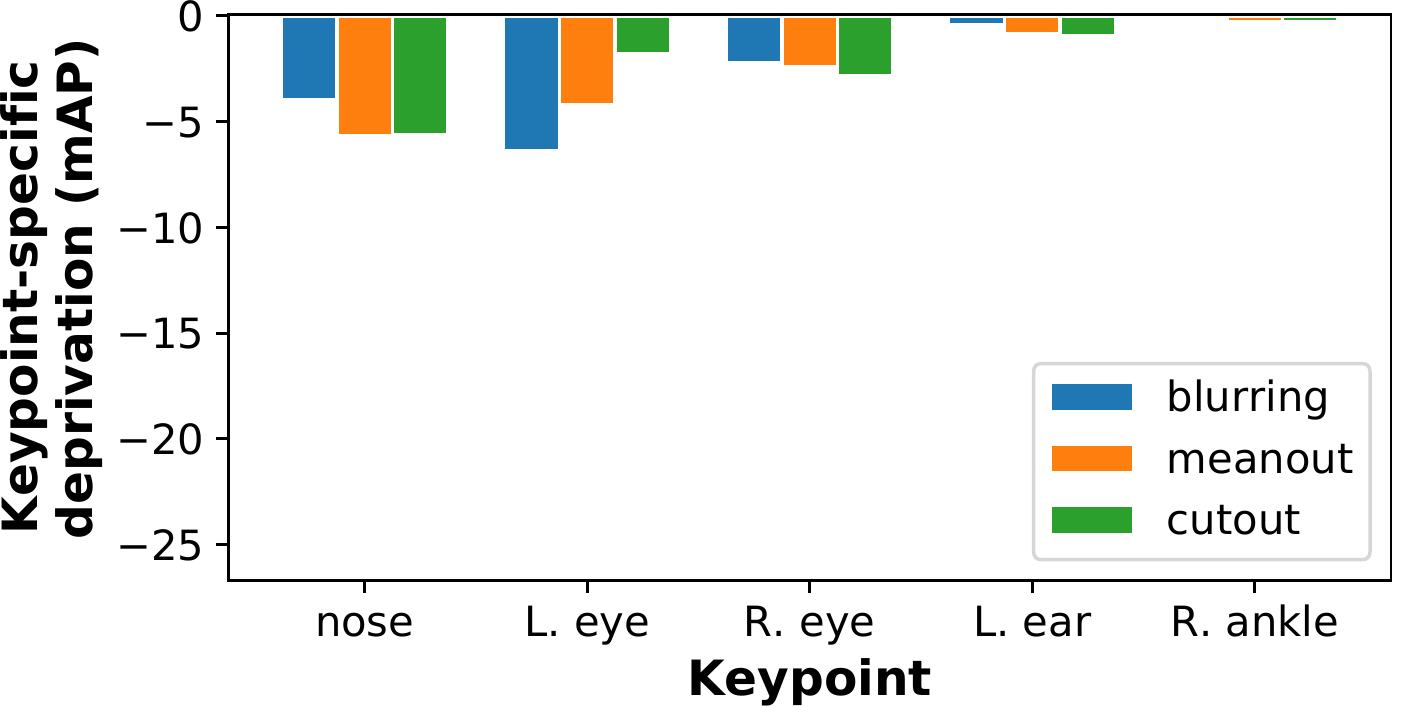}}
\hfill
\subcaptionbox{Left ankle is one of the least influential keypoints with loss only visible for meanout for occluded keypoint.} {\includegraphics[height=.125\textheight]{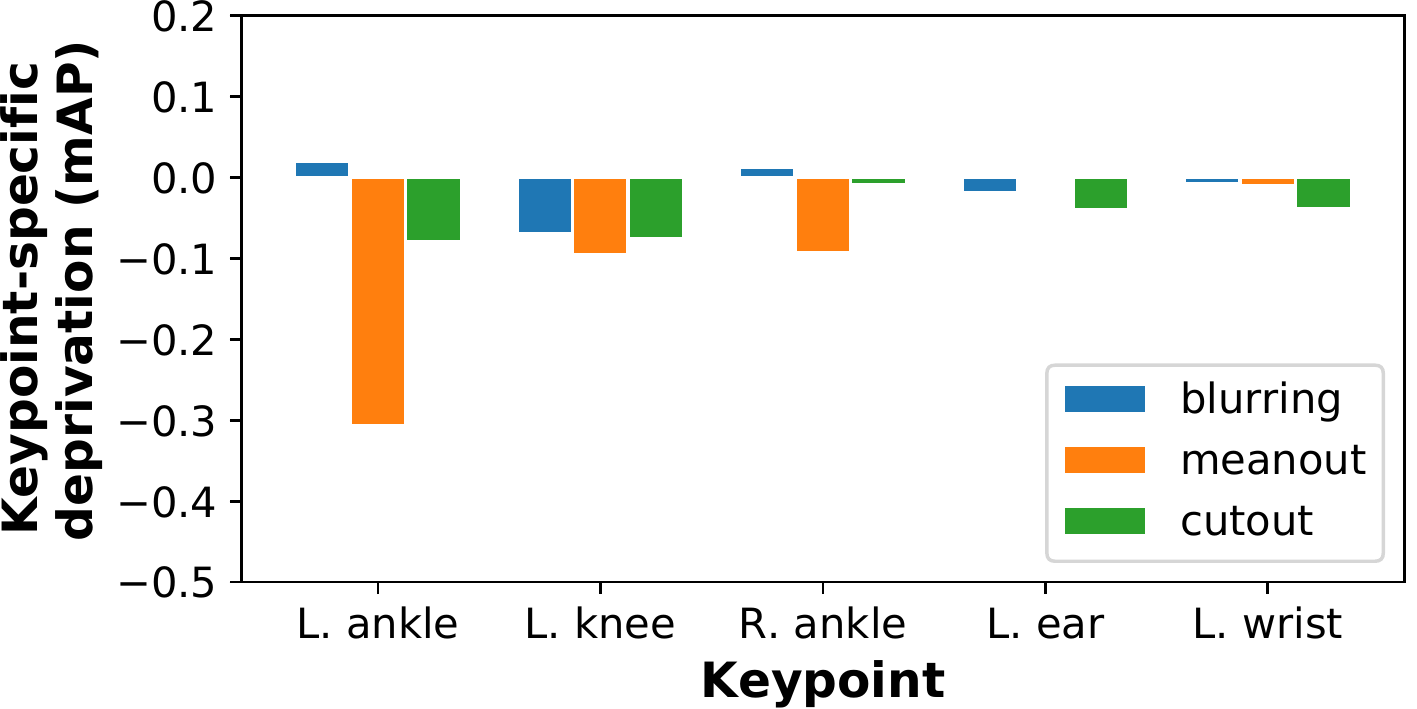}}
\caption{Loss in AP for top 5 keypoints with largest deprivation, when an individual key point is occluded.}
\end{figure*}

If we check the analysis of the reduced accuracy per keypoint for the case of attacking nose (Figure \ref{fig:keypoint-occlusions-nose}), the most affected keypoints are the ones within close distance, which are eyes and ears due to being a part of the head. Interestingly, occluding nose affects the performance of the left eye estimation more than occluding the left eye itself, respectively by approximately $10 \%$ and $5 \%$ (Figure \ref{fig:keypoint-occlusions-nose}, \ref{fig:keypoint-occlusions-left-eye}).
If we investigate per keypoint performance for occluding left ankle, it can be seen that the deprivation is by several magnitudes smaller than in case of the nose or left eye occlusions.
From the observation of the analyses, it can be drawn that HRNet \cite{HRNet} is not robust to keypoint occlusion attacks.

\subsection{How sensitive to part occlusion attacks?}



\begin{figure*}[!ht]
    \centering
    \includegraphics[height = 0.125 \textheight] {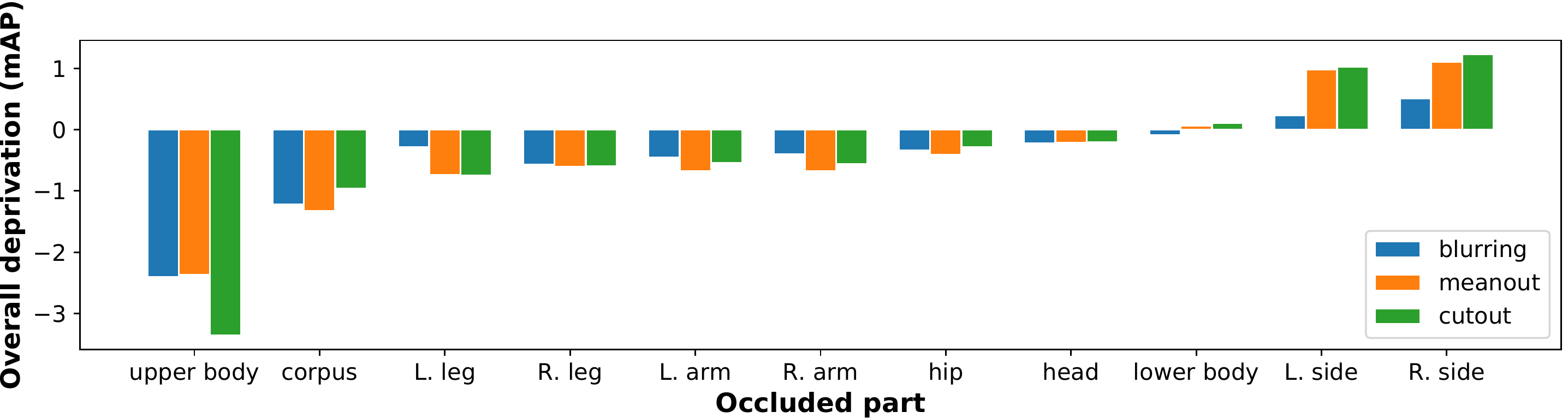}
    \caption{Change in mAP for various parts occluded. Upper body and sides are the parts that cause the highest loss in the performance.}
    \label{fig:part-occlusion}
\end{figure*}

\begin{figure*}[ht!]
\centering
\subcaptionbox{Significant loss in performance for all of the remaining keypoints. Blackout affects the method most. \label{fig:part-occlusion-upperbody}
}{\includegraphics[height=.125\textheight]{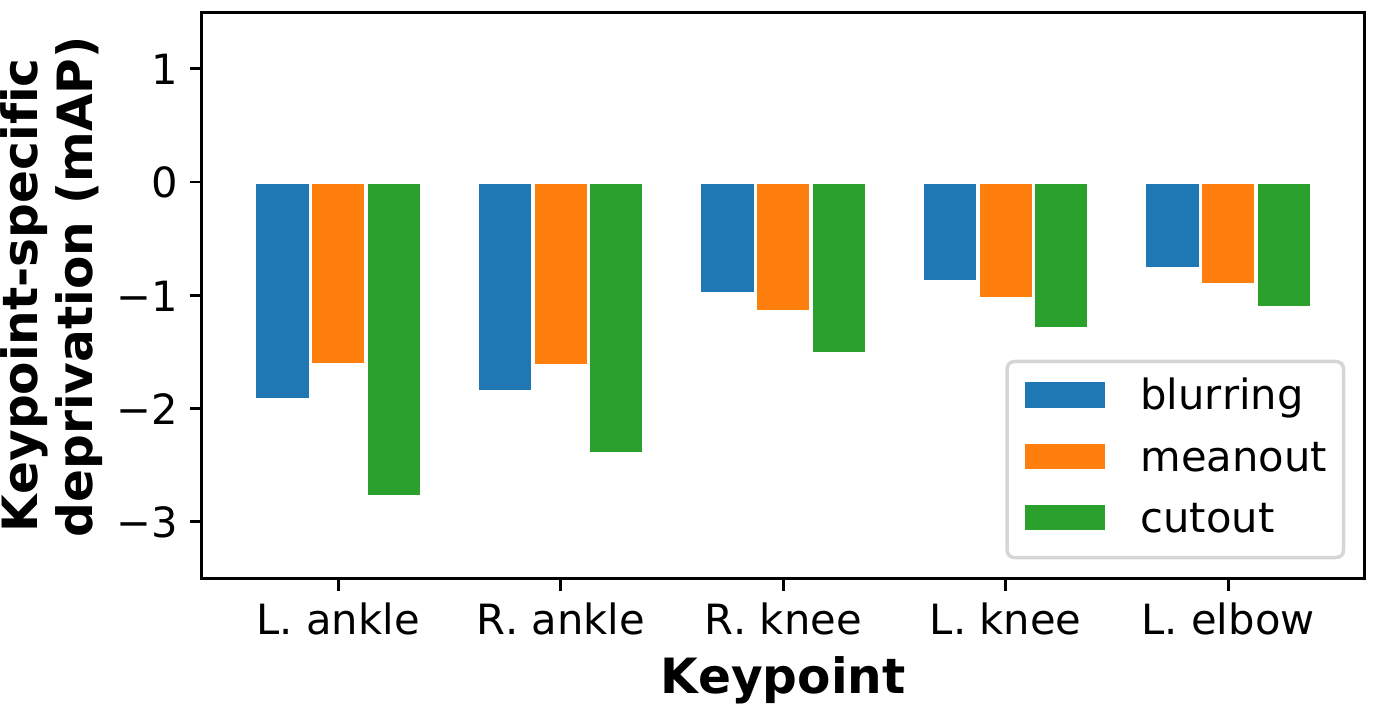}}
\hfill
\subcaptionbox{Similar loss across remaining keypoints, indicating that corpus is one of the most influential parts.         \label{fig:part-occlusion-corpus}
}{\includegraphics[height=.125\textheight]{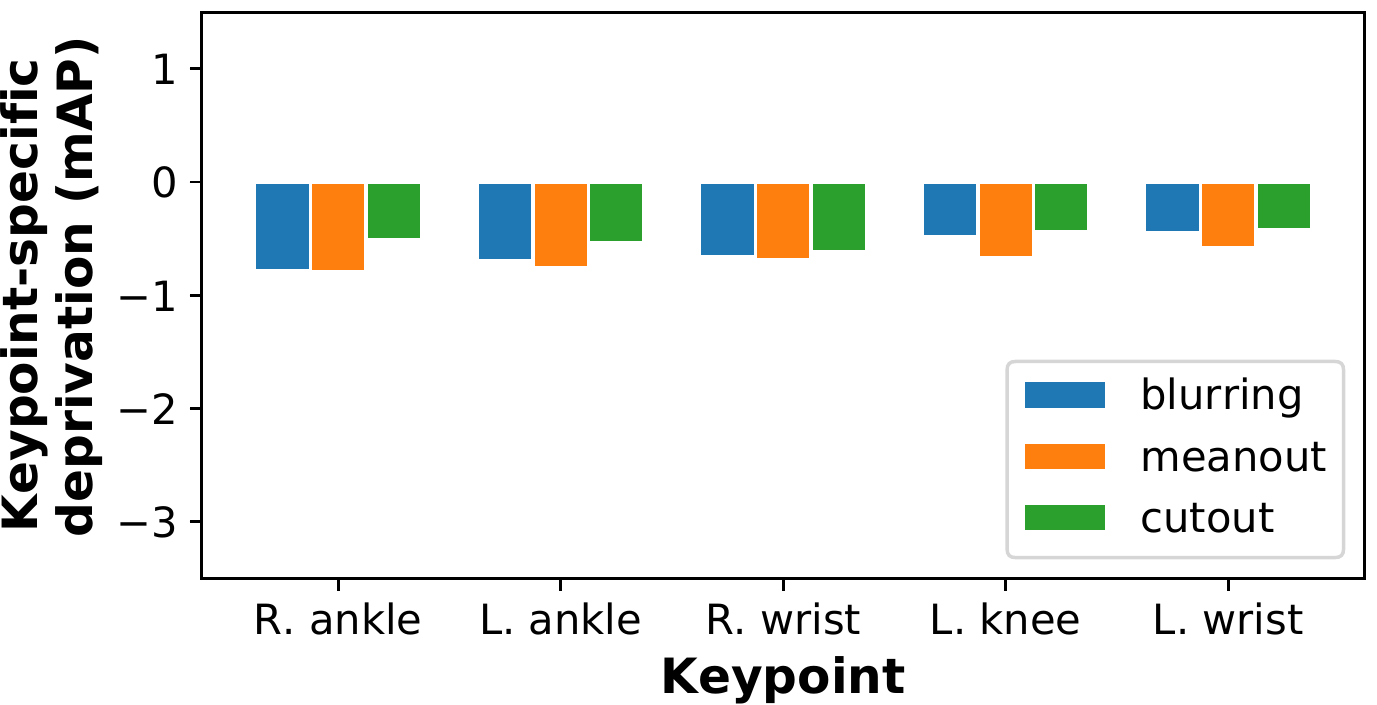}}
\hfill
\subcaptionbox{Occluding the left side of the body improves the performance of right shoulder, ear and elbow.        \label{fig:part-occlusion-rightside}
}{\includegraphics[height=.125\textheight]{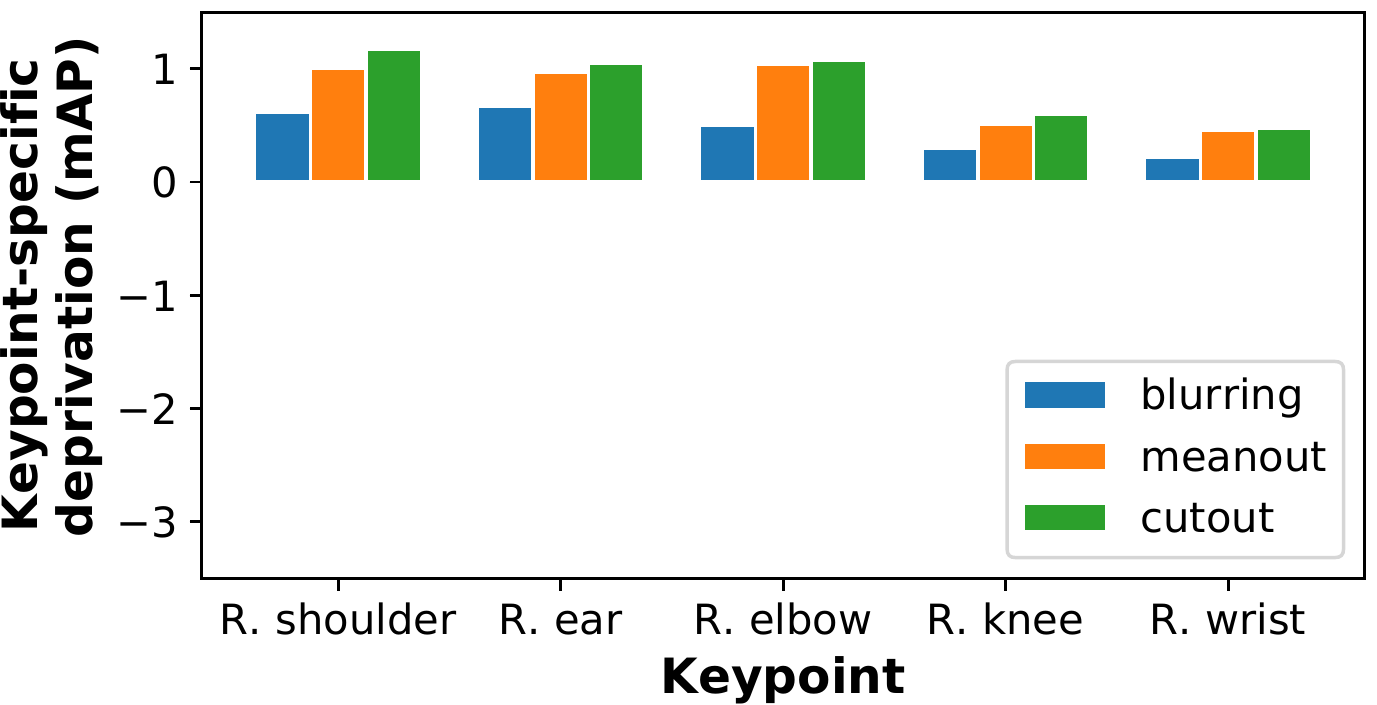}}
\caption{Change in AP for top 5 keypoints with the largest difference, when chosen part is occluded.}
\end{figure*}

We analyze the effect of the part occlusion attacks on each body parts given in Figure~\ref{fig:woodman}. Attacking the upper body, left and right sides influence the overall performance the most, by more than $44\%$, $24\%$ and $24\%$ with blackout attack respectively since these three parts include the majority of the keypoints (Figure~\ref{fig:part-occlusion}). When we examine keypoint-specific accuracy drops for the remaining keypoints of the upper body, it is clear that blackout is the most influential attack, with a drop of almost $3\%$ for left and right ankle (Figure~\ref{fig:part-occlusion-upperbody}).  If we investigate per-keypoint behaviour for the corpus (Figure~\ref{fig:part-occlusion-corpus}), we observe significant degradation of the  performance on all the keypoints, with left and right ankle affected the most.
Interestingly, attacking on one side improves performance of the the other side (Figure~\ref{fig:part-occlusion-rightside}). Attacking on left side increases the mAP score of right side such as shoulder, ear, elbow keypoints. The analysis demonstrates that the pose estimator is sensitive to part occlusion attacks.
\begin{figure*}[ht!]
\centering
\subcaptionbox{Blurring on left hip (almost not visible). \label{keypoint-targeted-blurring}}{\includegraphics[height=.15\textheight]{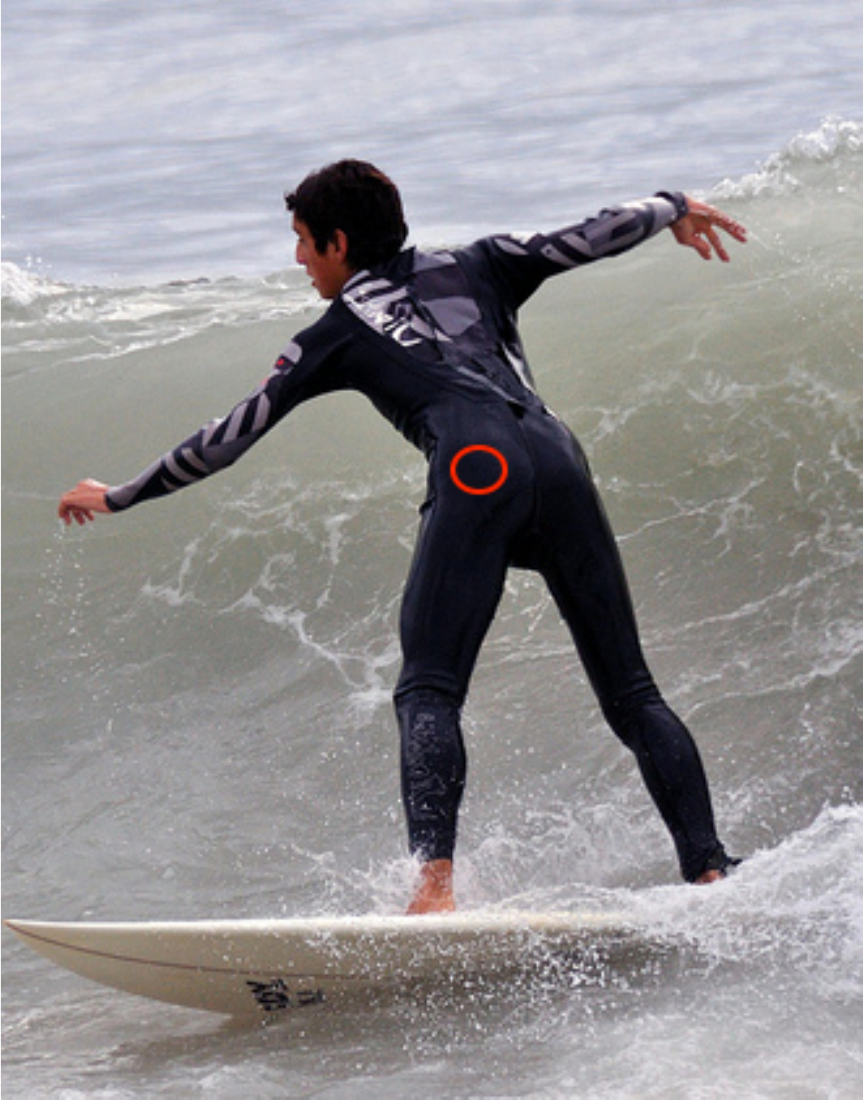}}%
\hfill 
\subcaptionbox{Cutout on left hip. \label{keypoint-targeted-cutout}}{\includegraphics[height=.15\textheight]{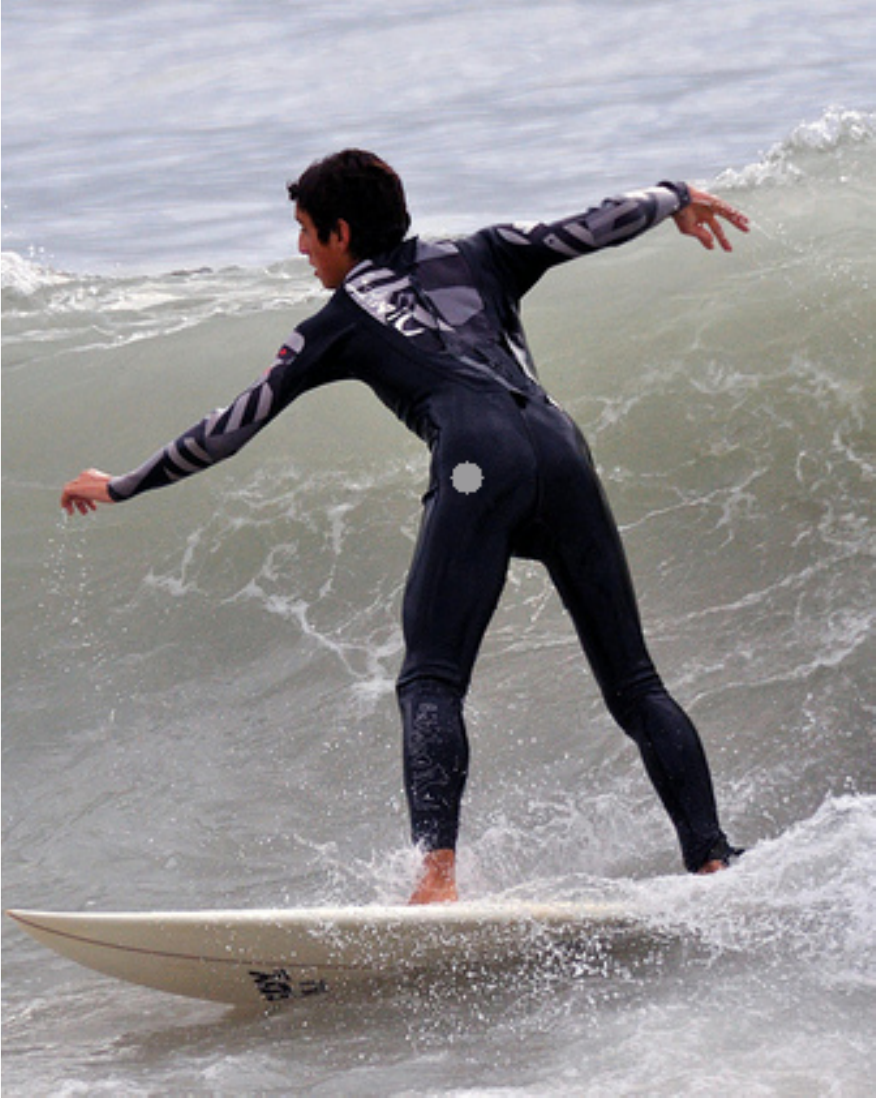}}%
\hfill
\subcaptionbox{Multi keypoint Cutout. \label{multi-keypoint}}{\includegraphics[height=.15\textheight]{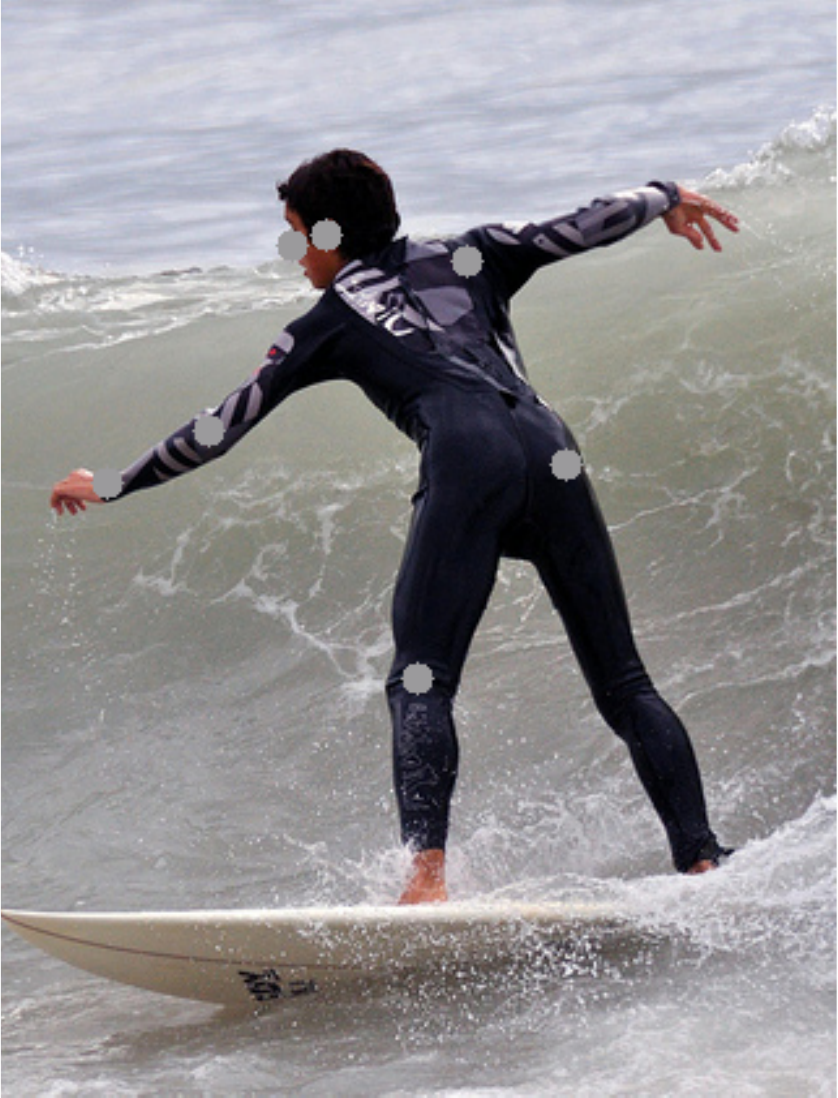}}%
\hfill
\subcaptionbox{Part Blurring. \label{part-targeted-blurring}}{\includegraphics[height=.15\textheight]{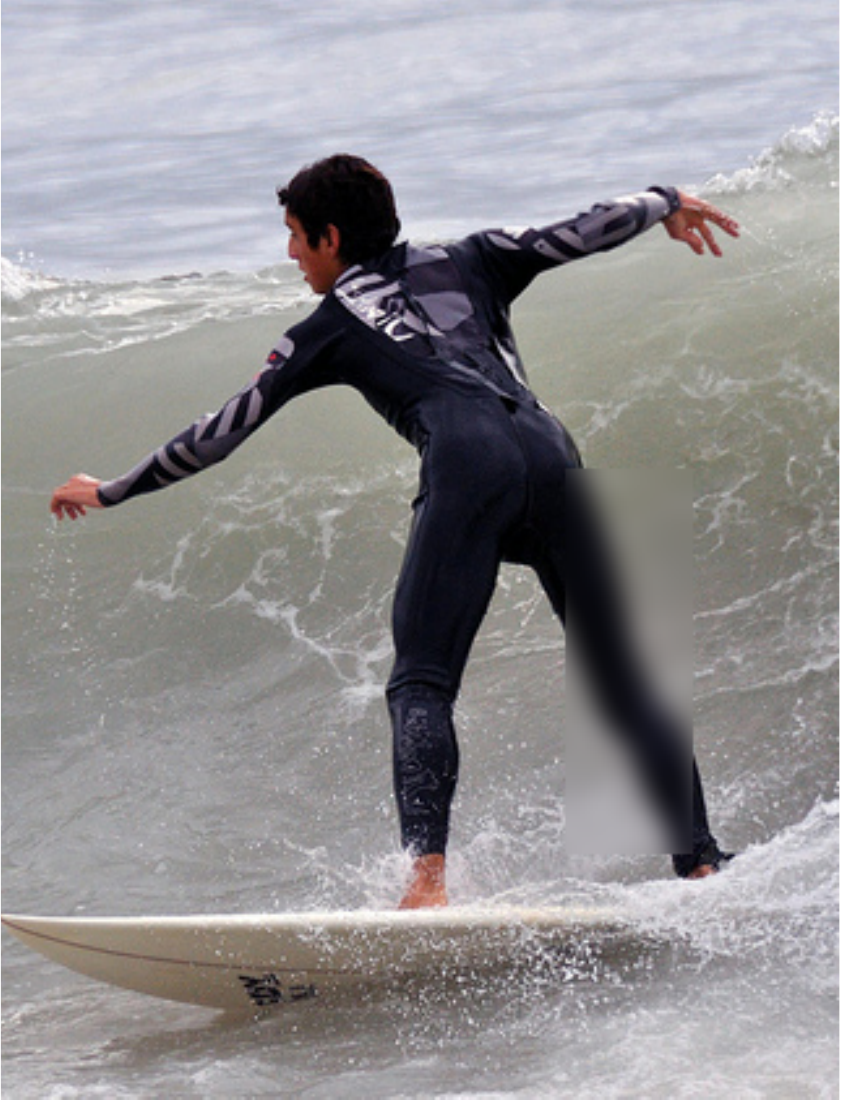}}%
\hfill
\subcaptionbox{Part Cutout.\label{part-targeted-cutout}}{\includegraphics[height=.15\textheight]{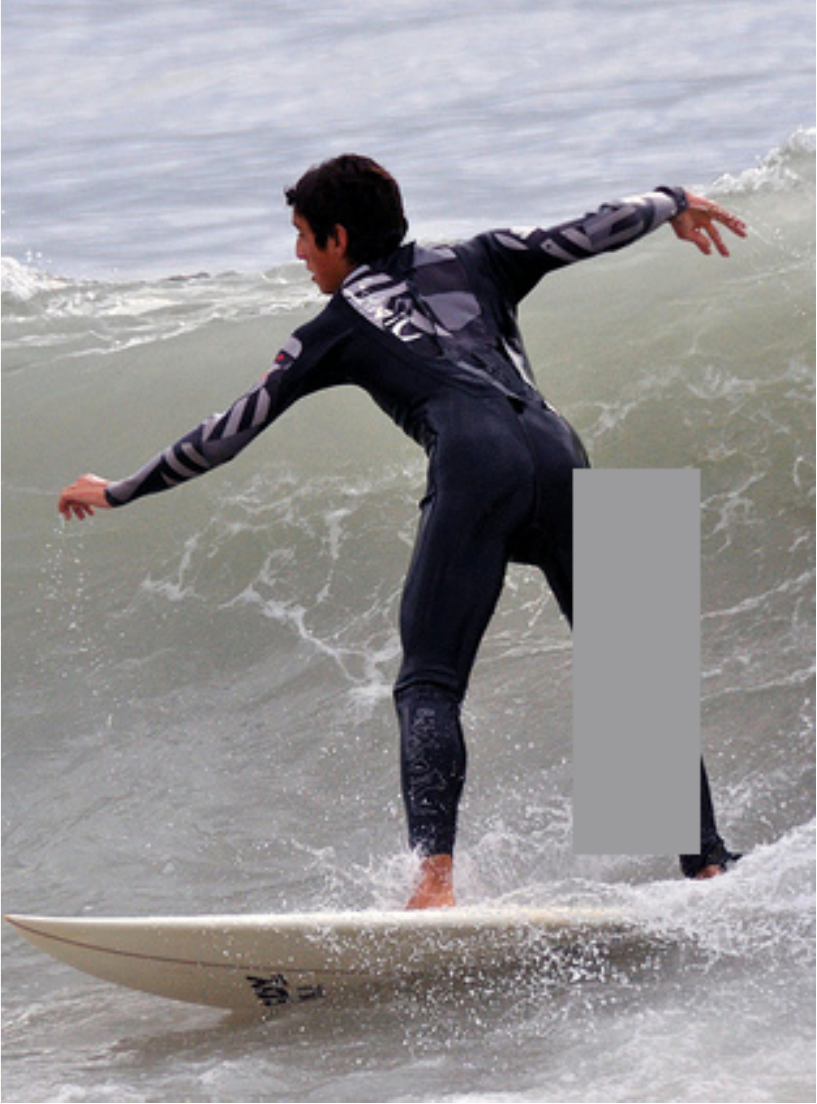}}%
\hfill
\subcaptionbox{PartMix on the right leg with a pasted random arm. \label{cutmix}}{\includegraphics[height=.15\textheight]{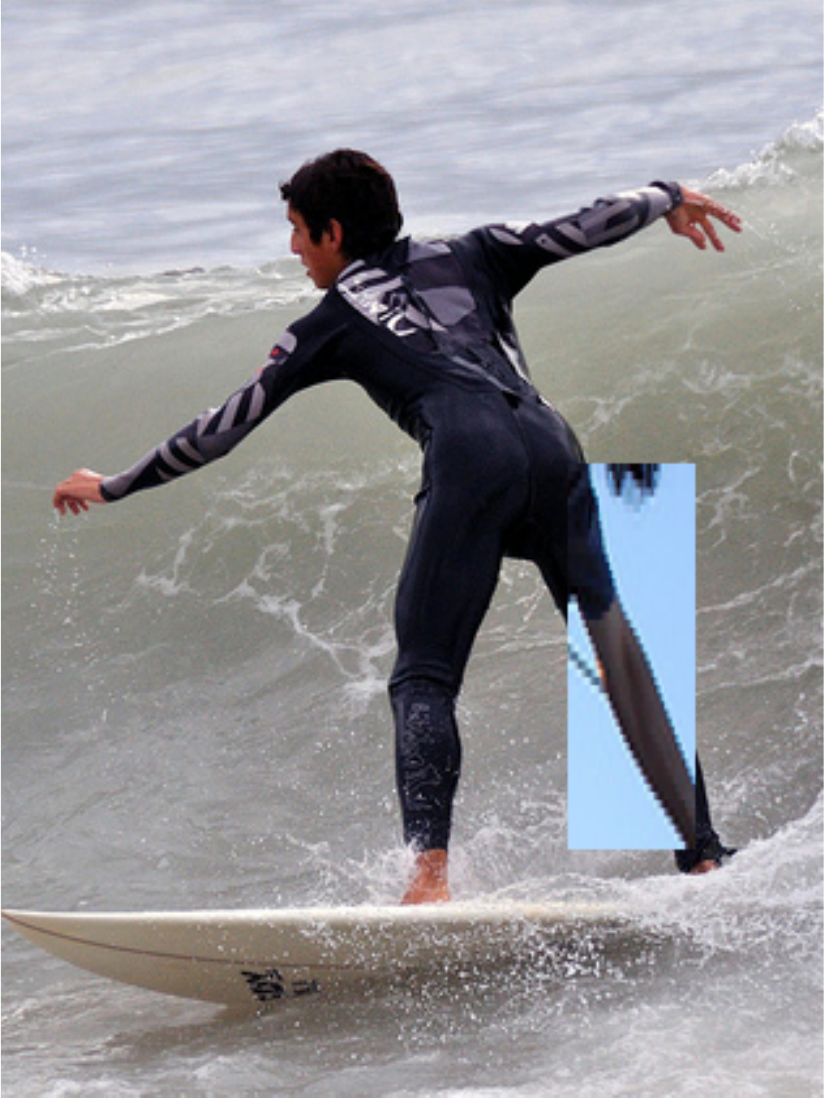}}%
\hfill
\caption{Targeted keypoint augmentations: a, b, c and targeted part augmentations: d, e, f.}
\end{figure*}

\section{Occlusion augmentation against attacks}


We evaluate two main human pose estimation datasets: COCO \cite{COCO_dataset} has 200k images with 250k person instances, labelled with 17 keypoints and 
MPII \cite{mpii} has 40k persons, each labelled with 16 joints. The train, validation and test sets include 22k, 3k and 15k person instances respectively. For the evaluation of MPII dataset, the validation set is used since the labels of the test set are not available.

For training HRNet~\cite{HRNet} models on COCO \cite{COCO_dataset} and MPII \cite{mpii} we follow the original pipeline of HRNet. For COCO dataset, human detection boxes are extended to fit 4:3 aspect ratio, and cropped from the image and resized to $256 x 192$. 
The pose estimator is trained with the keypoint location of the joints. 
The data augmentations that are used in HRNet training include random rotation $\in [-45^o,45^o]$, random scale $\in [0.65, 1.35]$, random flipping and half-body augmentations. The Adam optimizer \cite{adam} is used to train the network with the learning rate schedule following \cite{simple-baseline}, starting with $1e-3$ and reduced to $1e-4$ and $1e-5$ at 170th and 200th epochs respectively and the training is completed at the 210th epoch.
For MPII dataset, the training procedure of HRNet is as followed: 256x256 input size is used and half-body augmentations are discarded.
For the evaluation of the models, Object Keypoint Similarity (OKS) for COCO and Percentage of Correct Keypoints
(PCK) for MPII are used. 

During testing, HRNet firstly employs an object detection algorithm to obtain boxes with a single person. Afterwards the pose estimator produces the keypoint location of the joints. 
\subsection{Occlusion augmentation}


We investigate the following three methods:
(i) Targeted Blurring,
(ii) Targeted Cutout,
(iii) Targeted PartMix.
The augmentation techniques are called as \textit{targeted}, because we apply them on target locations of keypoints or parts instead of random location in the image. It is important to state that the proposed augmentation techniques are introduced after the bounding box person detection, and it thus does not affect the object detection method. 

\textbf{Targeted Blurring.} 
We use Gaussian blur for two types of targeted blurring: (i) keypoint blurring with a kernel size of 9 pixels (Figure~\ref{keypoint-targeted-blurring}) and (ii) part blurring with a kernel size of 31 pixels shown in Figure~\ref{part-targeted-blurring}. 

\textbf{Targeted Cutout.} The size of the keypoint cutout (Figure~\ref{keypoint-targeted-cutout}-\ref{multi-keypoint}) and part cutout (Figure~\ref{part-targeted-cutout}) are similar to the blurring equivalents. Instead of blurring, the area is colored with mean value of the image.

\textbf{Targeted PartMix.}
The method is designed to mitigate the occlusions caused by another person (Figure \ref{cutmix}).
In this approach, a different part from a random image is pasted in the place of a body part area. In this process, the keypoint labels of 
newly pasted part are not introduced to heatmap labels. 
This augmentation is only performed on body parts. Similar to the part level blurring and cutout augmentation methods, the occluded keypoints under the pasted area are still predicted.


\begin{table*}[t]
  \centering
  \begin{tabular}{l c c c c c c c c c}

    & \multicolumn{7}{r}{\small{\textbf{Evaluation results}}} \\
    \cmidrule(r){5-10}
    {\small\textit{Augmentation}}
    & {\small \textit{level}}
    & {\small \textit{removal}}
    & {\small \textit{p}}
    & {\small \textit{$AP$}}
    & {\small \textit{$AP^{50}$}}
    & {\small \textit{$AP^{75}$}}
    & {\small \textit{$AP^M$}}
    & {\small \textit{$AP^L$}}
    & {\small \textit{$AR$}}\\
 \midrule
 \midrule
 Baseline (no augments) & - &- &- & 65.3 & 86.4 & 72.6 & 62.6 & 70.7 & 70.2\\
 Baseline (flip, rot, scale) & - & - & - & 73.9 & 90.0 & 80.9 & 70.4 & 80.3 & 78.3\\
 Baseline (flip, rot, scale, half-body) & - & - & - & 74.3 & 90.6 & 81.7 & 70.7 & 80.7 & 78.8\\
 \midrule
\multirow{2}{*}{Blurring} & k & \xmark & 0.2 & 74.3 & 90.4 & 81.6 & 70.8 & 80.6 & 78.7 \\
  & k & \xmark  & 0.5 & 74.5 & 90.4 & 81.8 & 70.8 & 80.8 & 78.7 \\
\midrule
 \multirow{2}{*}{Cutout} & k & \xmark & 0.2 & 74.3 & 90.4 & 81.7 & 71.0 & 80.3 & 78.7 \\
 & k & \xmark  & 0.5  & 74.5 & 90.5 & 81.7 & 70.9 & 80.7 & 78.8 \\
\midrule
 \multirow{2}{*}{Cutout + Blurring} & k & \xmark  & 0.2 & 74.0 & 90.4 & 81.1 & 70.4 & 80.3 & 78.4 \\
& k & \xmark  & 0.5 & 74.3 & 90.5 & 81.1 & 70.8 & 80.6 & 78.6 \\
\midrule
\midrule
\multirow{3}{*}{Blurring}  & p & \checkmark & 0.2 & 74.3 & 90.5 & 81.7 & 70.6 & 80.8 & 78.6 \\
& p & \checkmark & 0.5 & 74.0 & 90.5 & 81.1 & 70.5 & 80.4 & 78.4 \\
 & p & \xmark & 0.5 & 74.1 & 90.3 & 81.1 & 70.6 & 80.2 & 78.5 \\
\midrule
\multirow{3}{*}{Cutout} & p & \checkmark & 0.2  & 74.2 & 90.5 & 81.2 & 70.8 & 80.4 & 78.6 \\
 & p & \checkmark & 0.5 & 74.2 & 90.3 & 81.1 & 70.6 & 80.4 & 78.6 \\
 & p & \xmark & 0.5  & 74.5 & 90.5 & 81.6 & 70.9 & 80.7 & 78.8 \\
 \midrule
\multirow{3}{*}{Cutout + Blurring} & p & \checkmark & 0.2 & 73.4 & 90.3 & 80.8 & 69.9 & 79.5 & 77.8 \\
 & p & \checkmark & 0.5 & 73.9 & 90.4 & 81.0 & 70.5 & 80.0 & 78.3 \\
 & p & \xmark & 0.5 & 74.3 & 90.4 & 81.2 & 70.6 & 80.5 & 78.6 \\
\midrule
Multikeypoint (max. 5)  & - & - &0.2 & 73.9 & 90.1 & 80.9 & 70.5 & 80.2 & 78.3 \\
\midrule
\multirow{2}{*}{PartMix}  & - & \checkmark & 0.5  & 74.3 & 90.5 & 81.1 & 70.7 & 80.6 & 78.7 \\
& - & \xmark & 0.5 & 74.4 & 90.7 & 81.5 & 71.1 & 80.5 & 78.8 \\
  \end{tabular}
  \caption{Comparison of augmentation variants on COCO validation set for HRNet using CascadeRCNN bounding boxes. Upper-part indicates single-keypoint augmentation and bottom-part shows multiple-keypoint augmentation. k and p in the level column represent keypoint and part augmentations respectively. Removal column indicates if the occluded keypoints are removed from prediction. Column $p$ is the probability of augmentation. Keypoint cutout and blurring, and part cutout and PartMix improve the performance. Other variants obtain results either on a par with baseline or worse than baseline.}
  \label{tab:abloation}
\end{table*}

\subsection{Analyses of occlusion augmentation}

All the following augmentation methods, except baselines, already include flipping, rotation, scaling and half-body augmentations. Each network obtains the boxes from Cascade RCNN~\cite{cascade_rcnn} detector which has ResNet50 backbone. The results of each method can be seen in Table \ref{tab:abloation}.

\textbf{Baselines.} Table \ref{tab:abloation} indicates 3 baseline variants. Firstly, HRNet without any augmentations obtains only $65.3\%$ mAP score. Secondly, adding flipping, rotation and scaling augmentations improve non-augmented baseline by $8.6\%$. Last variant is half body augmentation which adds only $0.4\%$ improvements on rotation and scaling augmentations. 

\textbf{Single keypoint augmentations.} We check the performance of 3 different augmentations: blurring, cutout and a combination of two of them which are applied on a single keypoint with the varying probability of 0.2 and 0.5 (Figure~\ref{keypoint-targeted-blurring}-\ref{keypoint-targeted-cutout}). 
We observe the highest improvement for blurring and cutout by $0.2\%$ when the probability is chosen as 0.5 (Table \ref{tab:abloation}). Other single keypoint variants do not improve the performance.

\textbf{Multi-keypoint augmentations.} We applied random multi-keypoint variant blurring and cutout with a maximum of 5 keypoints with a probability of 0.2 (Figure~\ref{multi-keypoint}). The augmentation decreases the model performance by $0.4\%$.

\textbf{Part augmentations.}
4 different part augmentation methods are used: part blurring, part cutout, a combination of both them and PartMix (Figure~\ref{part-targeted-blurring}, \ref{part-targeted-cutout} and \ref{cutmix} respectively).
To demonstrate the effect of each augmentation, we apply them with a probability of 0.2 and 0.5. In addition, the effect of removing the labels of the occluded keypoint is also investigated as \textit{removal} column in  Table \ref{tab:abloation}.

In the bottom part of Table
\ref{tab:abloation}, cutout and PartMix show $0.2\%$ and $0.1\%$ improvements respectively. In all the variants of blurring, small degradation or no improvement is observed. The combination of part level variants of cutout and blurring indicate some decreases of the performance for the removal configuration with probability of 0.2 and 0.5 and do not improve in non-removal configuration. 

To conclude to findings from the Table \ref{tab:abloation}, flipping, rotation and scaling augmentations add a huge performance gain to the HRNet. However, including half-body, the occlusion based augmentation methods  do not improve the performance of the pose estimator significantly.



\begin{figure}[ht!]
\centering
\includegraphics[width=.47\textwidth]{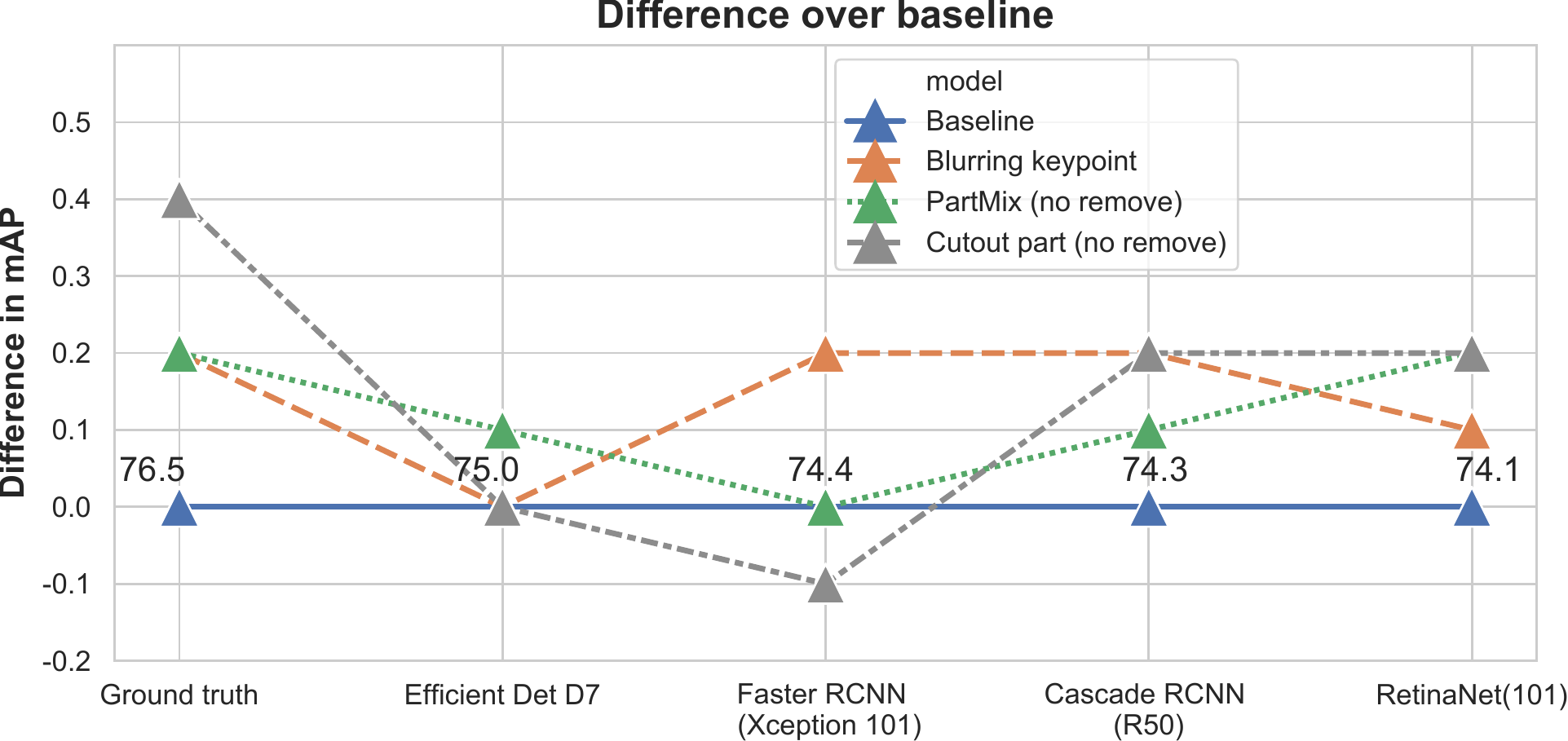}
\caption{Performance of chosen augmentations for HRNet-32 on various detection backbones and ground truth boxes. The ground truth bounding box performs best. Yet, none of the data augmentation methods help to improve performance over 0.2\% for any object detector. }
\label{fig:detector_performance}
\end{figure}


\begin{table*}[t]
  \centering
  \begin{tabular}{l c c c c c c c c c c c}
    & \multicolumn{7}{r}{\small{\textbf{Evaluation results}}} \\
    \cmidrule(r){5-12}
    {\small\textit{Augmentation}}
    & {\small \textit{level}}
    & {\small \textit{remove}}
    & {\small \textit{p}}
    & {\small \textit{Head}}
    & {\small \textit{Shoulder}}
    & {\small \textit{Elbow}}
    & {\small \textit{Wrist}}
    & {\small \textit{Hip}}
     & {\small \textit{Knee}}
    & {\small \textit{Ankle}}  & {\small \textit{\textbf{Total}}}\\
    \midrule \midrule
    {Baseline}&-& - & - & 97.1 & 95.9 & 90.4 &   86.4 & 89.1 & 87.2 & 83.3 & 90.3\\
    {Blurring} & k & \xmark &0.5 & 97.3 & 95.9 & 90.5 & 86.2 & 89.2 & 86.4 & 83.1 & 90.3 \\
    Cutout & p & \xmark& 0.5 & 97.2&  96.3 & 90.7 & 86.7 & 89.4 & 86.7 & 83.3 & \textbf{90.5}\\
    {PartMix} &- &\xmark & 0.5 & 97.4 & 96.2 & 91.0 & 86.8 & 89.2 & 86.7 & 83.0 & \textbf{90.5}\\
  \end{tabular}
  \caption{Results on MPII dataset. Keypoint blurring obtains on a par with the HRNet baseline, yet part cutout and PartMix increase the performance.}
  \label{tab:mpii_results}
\end{table*}


\textbf{The effect of the object detection algorithms.} HRNet \cite{HRNet} is a top-down approach which utilizes an object detection algorithm to obtain human instances. Therefore, the performance of the pose estimation considerably depends on the detection performance, namely detected human instances. 

By the evidence of the Table~\ref{tab:abloation}, we choose keypoint blurring, part cutout and PartMix methods for further analysis as they are the most promising augmentations.

We evaluate the pose estimation performances of vanilla HRNet and also of HRNet with the chosen augmentation methods with two 2-stage detectors, Faster RCNN \cite{faster_rcnn} with XCeption 101 backbone and Cascade RCNN \cite{cascade_rcnn}; 2 single-stage detectors, RetinaNet \cite{retinanet} and EfficientDet D7 \cite{efficientdet}; and by using ground truth boxes of human instances (Figure \ref{fig:detector_performance}). 

All the augmentations indicate improvements using ground truth bounding boxes by $0.2\%$  for keypoint blurring and PartMix, and $0.4\%$ for part cutout. All the chosen augmentation methods obtain better result with Cascade RCNN and RetinaNet $0.1-0.2\%$ depending on the augmentation. 
With EfficientDet D7 detector, keypoint blurring and part cutout result in similar to baseline except $0.1\%$ improvement by PartMix. 
For Faster-RCNN, keypoint blurring shows $0.2\%$ increase, yet part cutout degrades the performance by $0.1\%$. 

The performances of baseline and the augmentations vary depending on the object detector. The augmentation methods improves the results slightly, yet the gain is insignificant.

\textbf{Performance on MPII.}
We also test the data augmentation methods on MPII dataset (Table \ref{tab:mpii_results}).
If we check the total contribution of the proposed augmentations, keypoint blurring result in on a par with baseline, yet part cutout and PartMix increase the performance by $0.2\%$ for the metric PCK@0.5. 
The largest improvement per keypoint is observed for elbows by $0.6\%$ and wrists by $0.3\%$, with the degradation on knees and ankles by $0.4\%$ and $0.2\%$ respectively. 

Similar to analyses on the COCO dataset, the proposed augmentations can only improve the performance slightly.

\begin{figure}[ht!]
\centering
\subcaptionbox{}{\includegraphics[width=.49\columnwidth]{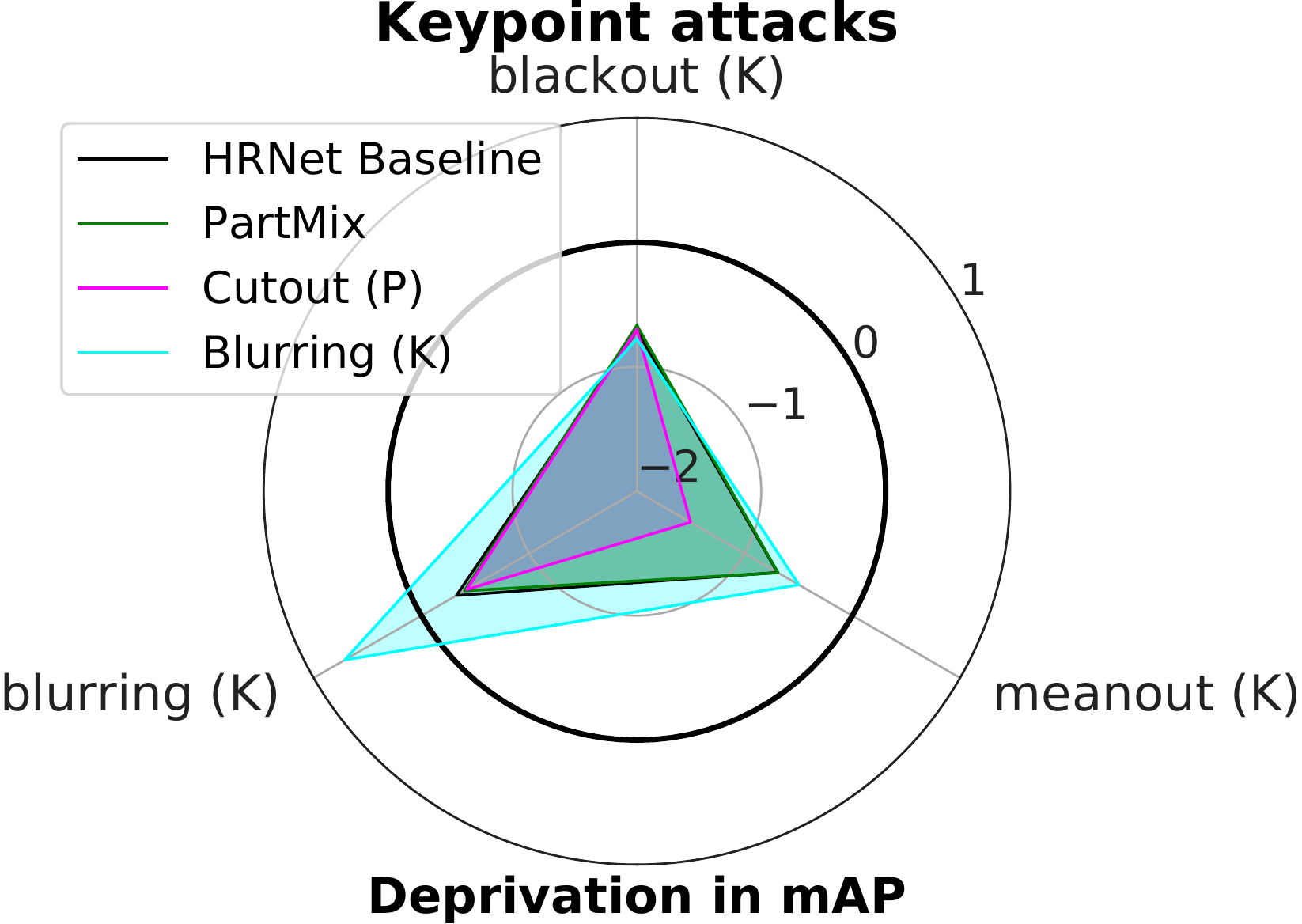}}
\subcaptionbox{}{\includegraphics[width=.49\columnwidth]{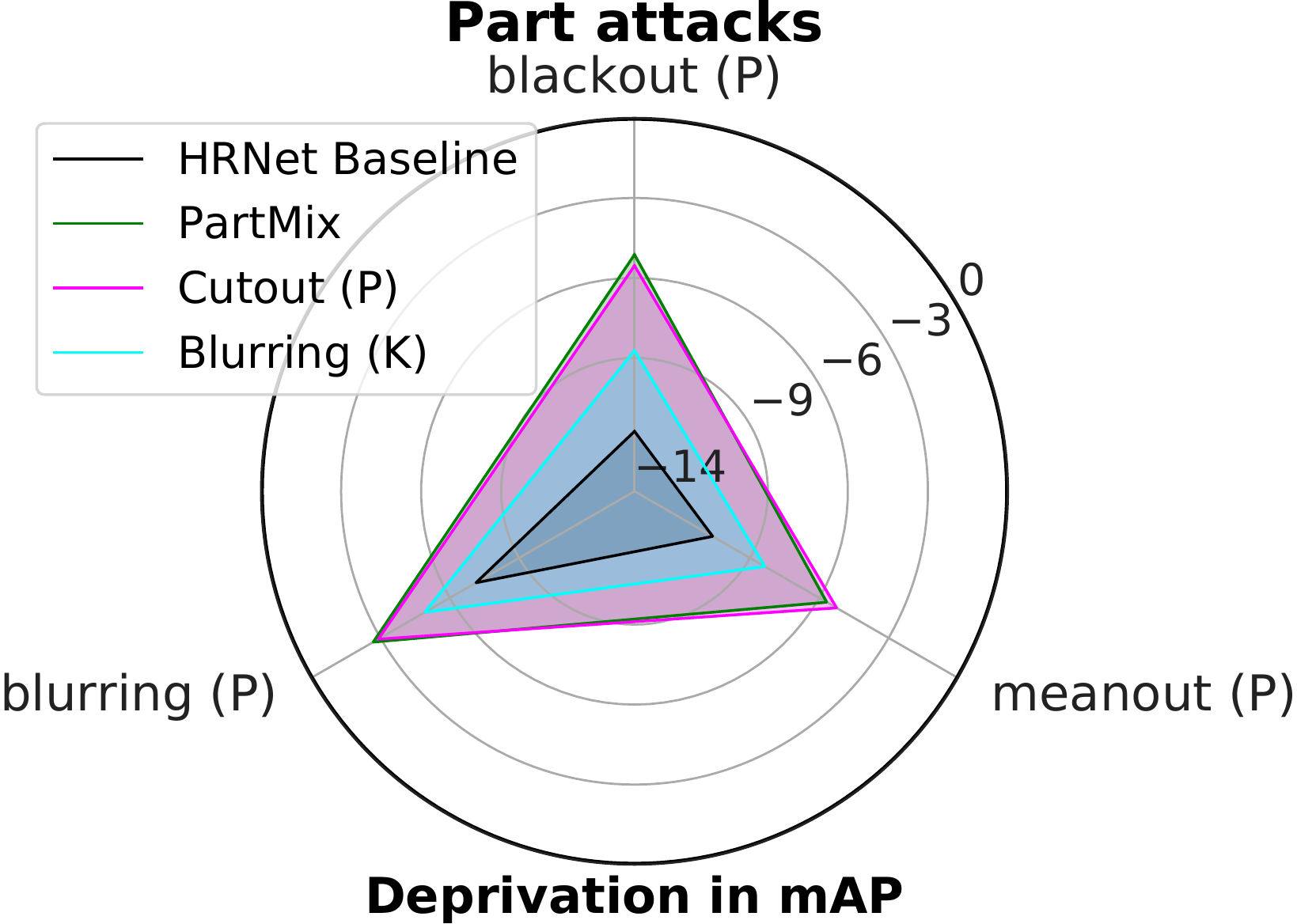}}
\caption{Robustness comparison  of proposed methods against (a) keypoint and (b) part occlusion attacks. 
Part augmentations improve the baseline but does not solve occlusion.}
\label{fig:ressilience_table}
\end{figure}


\textbf{How occlusion robust is data augmentation?}
Figure~\ref{fig:ressilience_table} shows the robustness of the baseline and the proposed augmentations to the occlusion attacks. The analysis is done on COCO dataset and the results are shown as mAP score of all keypoints.
We can clearly see that training with the keypoint blurring augmentation makes the network more robust against blurring attack, but there is no significant improvement for the other keypoint attacks.
In case of part attacks, we observe an improvement across all augmentation methods over the baseline.
For the part augmentations, there is a significant improvement against all part level attacks in comparison to baseline. 
Specifically, PartMix has almost no advantages against keypoint attacks, however, it improves part level methods about more than $5\%$ in comparison to baseline.
Part cutout obtains similar performance with PartMix against part attacks.
Proposed augmentations reduce the performance deprivations when we apply occlusion attacks, yet data augmentation still does not solve the occlusion problem.

\subsection{Augmentation on bottom-up method: Higher HRNet}

We also apply occlusion augmentations on Higher HRNet~\cite{higher_hrnet}, a bottom-up method. Higher HRNet is built on HRNet-32 and inputs 512x512 sized images. The training procedure follows Higher HRNet implementation from the paper. Unlike top-down methods, Higher HRNet operates on full-image and try to obtain the keypoints of each instance from the full-image. When applying the augmentations on Higher HRNet, we target all the human instances in the image.

Results in Table \ref{tab:higher_hrnet} show the augmentation methods to improve AP50 score slightly. For AP, all augmentations degrade performance by 0.6\% for keypoint level blurring, by 0.5\% for part level cutout and by 0.1\% for PartMix. Hence, using part and keypoint augmentations do not improve the performance of a bottom-up method.



\begin{table}[t]
  \centering
  \begin{tabular}{l c c c c c c c}
    & \multicolumn{4}{r}{\small{\textbf{Evaluation results}}} \\
    \cmidrule(r){2-6}
    {\small\textit{Augmentation}}
    & {\small \textit{$AP$}}
    & {\small \textit{$AP^{50}$}}
    & {\small \textit{$AP^{75}$}}
    & {\small \textit{$AP^M$}}
    & {\small \textit{$AP^L$}}\\
 \midrule
 \midrule
 Higher HRNet &67.1 & 86.2 & 73.0 & 61.5 & 76.1\\
 Blurring (K) & 66.5 & 86.3 & 72.1 & 60.6 & 75.7\\
 Cutout part (no remove) & 66.6 & 86.4 & 72.9 & 60.7 & 75.6 \\
 PartMix (no remove) & 67.0 & 86.4 & 73.0 & 61.3 & 75.8 \\
  \end{tabular}
  \caption{Results for bottom-up method, Higher HRNet~\cite{higher_hrnet}. The keypoint blurring, part cutout and Partmix degrade the performance of bottom-up methods. The augmentations do not help Higher HRNet.}
  \label{tab:higher_hrnet}
\end{table}
\section{Discussion and Conclusion}
In this study, we investigate the sensitivity of human pose estimators to occlusion. Firstly, we introduce targeted keypoint and body part occlusion attacks to show how much occlusion affects the performance. Secondly, keypoint and part based data augmentation techniques against occlusion are investigated. The structured analyses indicate that deep pose estimators are not robust to occlusion.
With all the bells and whistles, the current and proposed data augmentation methods do \textbf{not} bring significant improvements on the performance of the top-down pose estimators and even reduce the performance for the bottom-up approaches. Our paper is important because it helps data scientists looking for improvements against occlusions to not work on data augmentation. Battling occlusions is still an open problem for human pose estimation. 


Part based attacks and augmentation are applied as a rectangle shape. This fact can introduce unusual artefacts because natural occlusions can have arbitrary shapes. 
Each keypoint augmentation is applied as a circle that covers the related keypoint, yet in reality, keypoint occlusions can occur with numerous shapes and ways e.g. self occlusion, occlusion by other object.
Moreover, for bottom-up approaches, the input image into the network may have more perturbations since the full image can contain multiple instances. This fact can harm the learning process.






{\small
\bibliographystyle{ieee_fullname}
\bibliography{main}
}
\pagebreak
\appendix
\section*{More results on COCO val set}
\textbf{HRNet results.}
For this experiment, we increase the input resolution of images from 256x192 to 384x256. The training process follows the aforementioned scheme for COCO dataset.

\begin{figure}[!ht]
\centering
    \includegraphics[width = 0.5 \textwidth] {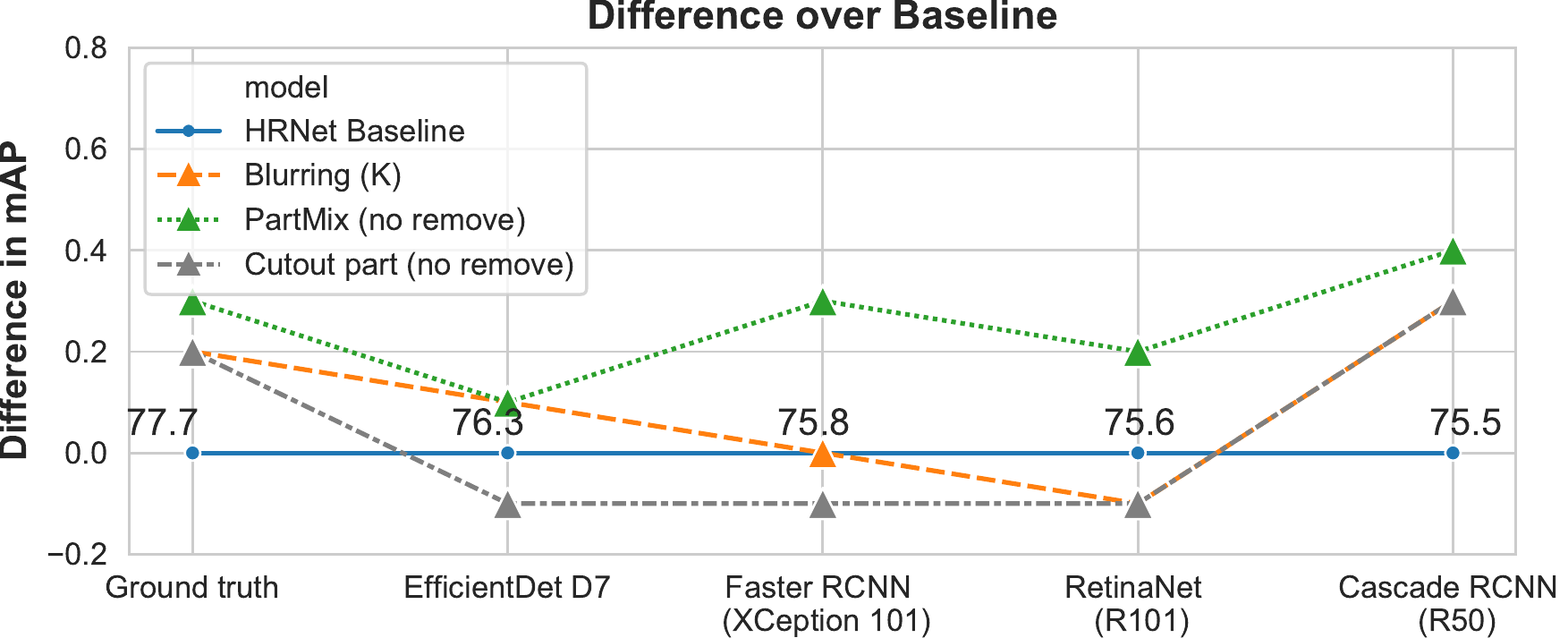}
    \caption{Higher resolution input for HRNet 32: the resolution is changed from 256x192 to 384x256. The best performance across detection backbones is observed for PartMix.}
    \label{fig:higher_res}
\end{figure}

According to the analysis of the performance across a variety of detection backbones shown in Figure \ref{fig:higher_res}, we notice that PartMix is consistently improving performance - with the greatest boost of 0.4\% for Cascade R-CNN and 0.3\% for Faster RCNN. For both keypoint blurring and part cutout, we observe no significant improvement or even the performance decreases - for part cutout using EfficientDet, Faster RCNN and RetinaNet and for Blurring using RetinaNet. All the presented augmentations show largest gain for Cascade RCNN.
Occlusion augmentations do not help to solve occlusion problems when higher resolution input is used.

\begin{figure}[!ht]
\centering
    \includegraphics[width = 0.5 \textwidth] {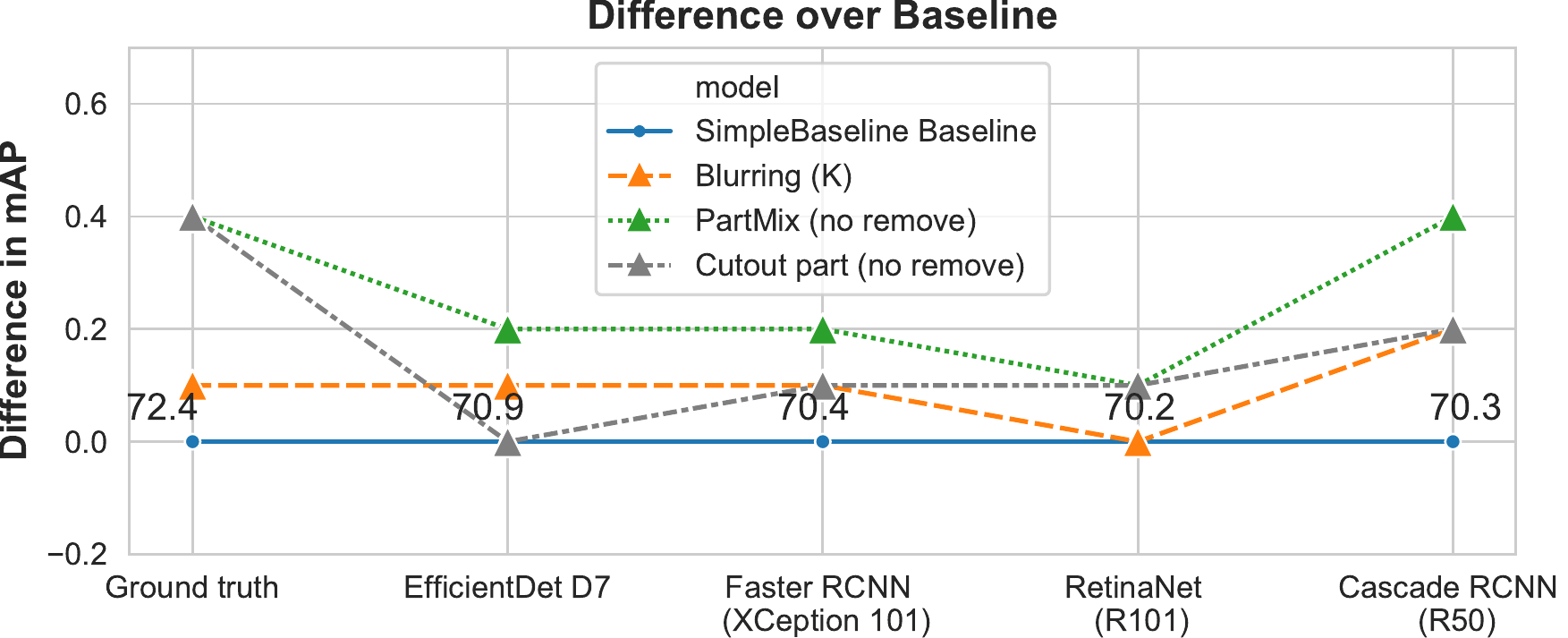}
    \caption{Performance of chosen augmentations for SimpleBaseline on various detection  backbones and ground truth boxes. Using the ground truth bounding boxes outperforms all the SimpleBaseline methods with a detection backbones.}
    \label{fig:resnet}
\end{figure}

\textbf{SimpleBaseline results.}
The usability of occlusion augmentations are not only limited to HRNet, yet they can be used with other top-down methods like SimpleBaseline~\cite{simple-baseline}. In this experiment, we apply the occlusion augmentations on SimpleBaseline method with different object detection backbones. The training procedure of the network follows the original implementation. 

By checking the performance across the various detection backbones we observe either small or no improvement at all (Figure \ref{fig:resnet}). PartMix show the most significant improvement across detection backbones, with 0.4\% boost in the performance for the ground truth boxes and the boxes produced by Cascade RCNN, 0.2\% for EfficientDet and Faster RCNN and 0.1 \% for RetinaNet. Cutout and Blurring improve at most 0.2\% across all the detection backbones, apart from 0.4\% for Cutout using ground truth bounding boxes. According to the results, proposed augmentation techniques do not solve occlusion problems of SimpleBaseline method.

\begin{figure}[!ht]
\centering
    \includegraphics[width = 0.35 \textwidth] {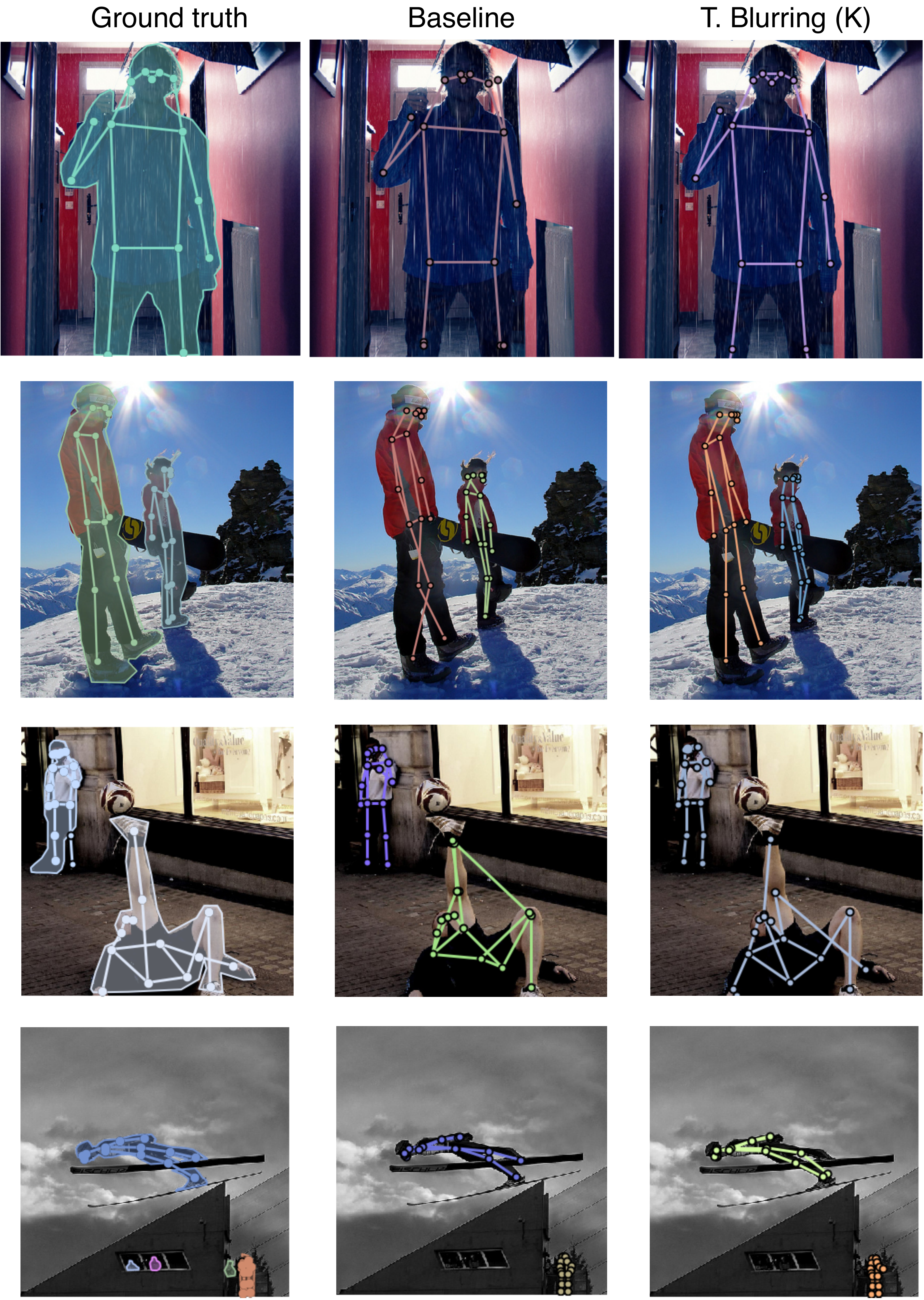}
    \caption{Qualitative comparison between ground truth (left), baseline (middle) and keypoint Blurring (K) (right). 
    1st and 2nd rows respectively - misplacement of left wrist keypoint and mismatch between knee keypoints in the baseline and keypoint blurring fixes the mistakes. 3rd row - both baseline and proposed method produce wrong keypoints. 4th row - baseline produces near-optimal keypoints whilst keypoint blurring makes mistake on left ankle keypoint. Data augmentation does not solve occlusion problem.}
    \label{fig:examples}
\end{figure}

\section*{Visualization of results}
Figure \ref{fig:examples} presents a qualitative comparison between ground truth, HRNet-32 Baseline and keypoint blurring augmentation. 
In the first and second rows, keypoint blurring outperforms the baseline by obtaining the position
of the left wrist and knee keypoints respectively. In the third row, both baseline and keypoint blurring produce wrong keypoint predictions. Fourth row presents failure case when baseline produces near-optimal annotations, while the method with keypoint blurring predicts left ankle in place of the right one.

\end{document}